\documentclass[runningheads]{llncs}

\usepackage{eccv}

\usepackage{eccvabbrv}

\usepackage[accsupp]{axessibility}  % Improves PDF readability for those with disabilities.
\usepackage{booktabs}
\usepackage{epsfig}
\usepackage{amssymb}
\usepackage{amsmath,graphicx}
\usepackage{multirow}
\usepackage{multicol}
\usepackage{siunitx}
\usepackage{tabularx}
\usepackage{xcolor}
\usepackage{comment}
\usepackage{bm}
\usepackage{tikz}
\usepackage{makecell}
\usepackage{pifont}% http://ctan.org/pkg/pifont
\usepackage{color}
\usepackage{fancyhdr}
\usepackage{arydshln}
\usepackage{verbatim}
\usepackage{footmisc}
\usepackage{float}

\usepackage[normalem]{ulem}

\usepackage{paralist, algorithm, algorithmic}
\usepackage{wrapfig}

\makeatletter
\renewcommand{\ALG@name}{\scriptsize Algorithm} 
\makeatother

\usepackage{hyperref}

\usepackage{orcidlink}

\begin{document}

\title{Gradient-Regularized Out-of-Distribution Detection} 

\titlerunning{Abbreviated paper title}

\author{Sina Sharifi\inst{1*}  \and
Taha Entesari\inst{1*}\and
Bardia Safaei\inst{1*}\and \\
Vishal M. Patel\inst{1} \and
Mahyar Fazlyab\inst{1}} 

\authorrunning{Sharifi et al.}

\institute{Johns Hopkins University, Baltimore MD 21218, USA\\
\email{\{sshari12, tentesa1, bsafaei1, vpatel36, mahyarfazlyab\}@jhu.edu}}

\maketitle
\def\thefootnote{*}\footnotetext{Co-first authors, equally contributed to the work}

\begin{abstract}
    One of the challenges for neural networks in real-life applications is the overconfident errors these models make when the data is not from the original training distribution.
    Addressing this issue is known as Out-of-Distribution (OOD) detection.
    Many state-of-the-art OOD methods employ an auxiliary dataset as a surrogate for OOD data during training to achieve improved performance. 
    However, these methods fail to fully exploit the local information embedded in the auxiliary dataset.
    In this work, we propose the idea of leveraging the information embedded in the gradient of the loss function during training to enable the network to not only learn a desired OOD score for each sample but also to exhibit similar behavior in a local neighborhood around each sample.
    We also develop a novel energy-based sampling method to allow the network to be exposed to more informative OOD samples during the training phase. This is especially important when the auxiliary dataset is large. We demonstrate the effectiveness of our method through extensive experiments on several OOD benchmarks, improving the existing state-of-the-art FPR95 by $4\%$ on our ImageNet experiment.
    We further provide a theoretical analysis through the lens of certified robustness and Lipschitz analysis to showcase the theoretical foundation of our work.
    Our code is available at \href{https://github.com/o4lc/Greg-OOD}{https://github.com/o4lc/Greg-OOD}.
    % We will publicly release our code after the review process.
\end{abstract}

\section{Introduction}
Neural networks are increasingly being utilized across a wide range of applications and fields, achieving unprecedented performance levels and surpassing traditional state-of-the-art approaches.
However, concerns about robustness and safety, coupled with significant challenges in verifying robustness and ensuring safe performance impede their use in more sensitive applications \cite{li2023data, Bafghi_2023_CVPR}. One concern that arises when deep models are deployed in the real world is their tendency to produce over-confident predictions upon encountering unfamiliar samples that are distant from the space in which they were trained \cite{nguyen2015deep, park2023nearest}. As a result, the field of Out-of-Distribution (OOD) detection has emerged to study and address this phenomenon.

\begin{figure}[t!]
    \centering
    \begin{subfigure}[t]{.45\textwidth}
        \includegraphics[width=\textwidth]{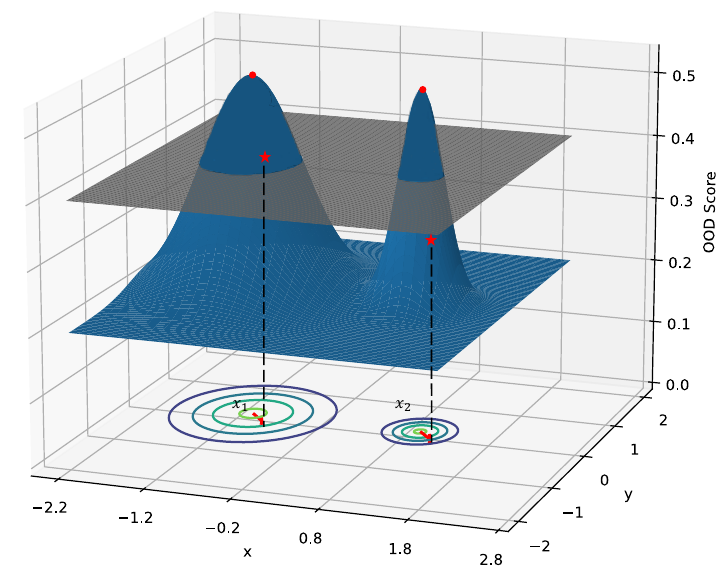}
        \caption{}
        \label{fig:Greg}
    \end{subfigure}
    \begin{subfigure}[t]{.45\textwidth}
        \includegraphics[width=\textwidth]{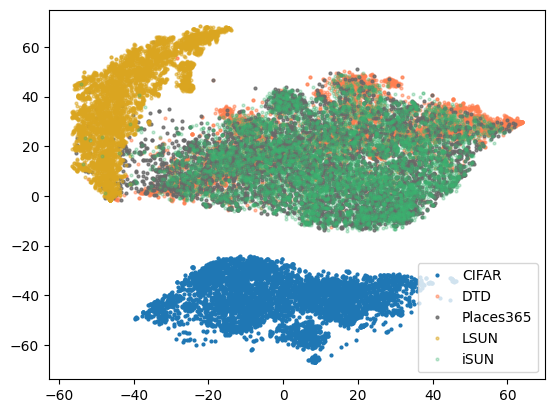}
        \caption{}
        \label{fig:tsne}
    \end{subfigure}
    \caption{\textit{Left:}
        Effect of the local structure of the score manifold on a two-dimensional toy example for OOD detection.
        The grey plane depicts the score function's decision threshold.
        An equal amount of perturbation in these two scenarios results in different OOD detections, highlighting the importance of the local structure of the score manifold.
        \textit{Right:} t-SNE plot showing the representation of ID and OOD datasets for CIFAR experiments.}
\end{figure}

Many works have been presented in recent years, trying different methods and ideas to regulate the network's output to OOD samples in terms of softmax probability \cite{hendrycks2017a, Zhang_2023_CVPR, noh2023simple} or energy metrics \cite{liu2020energy}.
One family of methods, known as post-hoc approaches \cite{peng2024conjnorm, lu2024learning, xu2023vra, yuan2024discriminability, ahn2023line, olber2023detection, tang2024cores, liang2017enhancing, liu2023gen}, operate on a given trained network and employ certain statistics, such as the confidence level of the network, the number of active neurons, the contribution of important pathways in the network, etc., to perform OOD detection without using any samples from an auxiliary distribution.
Other methods \cite{chen2021atom, liu2020energy, ming2022poem, jiang2023diverse, hendrycks2018deep} use OOD data for training or finetuning a network. The latter family usually tends to achieve better performances in comparison to post-hoc methods, as their access to the auxiliary dataset provides them with additional information regarding the outlier distribution. OOD detection is closely related to anomaly \cite{perini2024unsupervised, yang2024follow} and novelty detection \cite{lo2022adversarially, choi2023projection}, each with slightly different problem formulations. Our work focuses on the general OOD detection problem.

In this paper, we propose \emph{GReg}, a novel \textbf{G}radient \textbf{Reg}ularized OOD detection method, which aims to learn the local information of the score function to improve the OOD detection performance. 
Using information from the gradient of the score function, our method builds on top of current state-of-the-art OOD detection methods (that require training) to incorporate local information into the training process and lean the network towards a smoother manifold. 

To motivate this idea, consider \Cref{fig:Greg}. This figure shows a two-dimensional example of a possible scenario in OOD detection. In this example, $x_1$ and $x_2$ are two OOD samples and the network has learned to assign the same OOD scores  (denoted by red dots in the figure) to both samples and detect them correctly as OOD. 
However, the difference in the local behavior of the network at these points, leaves samples in the vicinity of $x_2$ susceptible to misdetection, whereas, samples close to $x_1$ are always detected as OOD.
To see this, consider the perturbations indicated by the red arrows, where the resulting points in each case represent two possible samples with equal distance from points $x_1$ and $x_2$, respectively.
However, as can be seen in the figure, the OOD detection algorithm cannot detect one of the samples correctly.
This emphasizes the importance of regulating the local behavior of deep learning models, especially when OOD robustness is crucial.
Furthermore, \Cref{fig:tsne} shows the t-SNE plot of the CIFAR and some of the well-known OOD benchmarks. This further confirms that even in high dimensions the ID data (CIFAR) are well-separated from the OOD data.

On top of that, several recent works have shown the effectiveness of some form of informative sampling or clustering on OOD detection \cite{ming2022poem, chen2021atom, jiang2023diverse}.
Sampling gains more importance specifically when a large dataset of outlier samples is available and the network cannot be exposed to all of the data during training.
Inspired by such works, we present \emph{GReg+}, the coupling of \emph{GReg} with an energy-based clustering method aimed at making better use of the auxiliary data during training.
This is in line with our intuition of utilizing local information as clustering enables the use of samples from diverse regions of the feature space.
The diversity put forward by the clustering mechanism allows \emph{GReg+} to achieve its state-of-the-art performance in more regions of the space rather than only performing well on regions well-represented in the dataset. 

\noindent In summary, our key contributions are:
 \begin{itemize}
     \item We propose the idea of regularizing the gradient of the OOD score function during the training (or fine-tuning). 
     This allows the network to better learn the local information embedded in ID and OOD samples.
     \item We propose a novel energy-based sampling method that chooses samples --based on their energy levels-- from the OOD dataset that represent more vulnerable regions of the space using clustering techniques. 
     \item We provide empirical results and ablation studies on a wide range of architectures and different datasets to showcase the effectiveness of our method, along with an extensive comparison to the state-of-the-art. 
     In addition, we provide a detailed theoretical analysis to justify our results.
 \end{itemize}

To the best of our knowledge, this is the first work that utilizes the norm of the gradient of the score function during the training phase to learn the local behavior of the ID/OOD data aiming to improve the OOD detection performance.

\section{Related Work}
Several works have attempted to enable deep neural networks with OOD detection capabilities in recent years 
\cite{zheng2023out, cao2024envisioning, li2023rethinking, ming2022delving, wang2023out, oza2019c2ae, Fan_2024_CVPR, safaei2023open, chen2020learning, zhang2023openood}.
These works can be divided into two main groups: methods that do not use auxiliary data (post-hoc) and those that utilize auxiliary data.
\subsection{Post-hoc OOD Detection}
    In one of the earliest attempts to remedy the OOD detection problem, \cite{hendrycks2017a} uses the maximum softmax probability (MSP) score to recognize OOD samples during inference. ODIN \cite{hendrycks2018deep} improves the MSP score's separability by adding small noise to the input and utilizing temperature scaling. \cite{liu2020energy, huang2021importance, lin2021mood, liu2023gen} aim to formulate enhanced scoring functions to better distinguish OOD samples from ID samples. For example in \cite{liu2020energy}, the authors propose using energy as a powerful OOD score. GradNorm \cite{huang2021importance} uses the vector norm of the gradient of the KL divergence between the softmax output and a uniform probability distribution to devise an OOD score. 
    It was later observed by \cite{sun2021react}, that OOD data tend to have different activation patterns. As a solution, they propose activation truncation to improve OOD detection.
    Building on this observation, DICE \cite{sun2022dice} ranks weights based on the measure of contribution, and selectively uses the most contributing weights. 
    LINe \cite{ahn2023line} further improves on DICE by using Shapley values \cite{shapley1953value} to more accurately detect important and contributing neurons.
    
\subsection{OOD Detection with Auxiliary Outlier Dataset}
    When an auxiliary dataset is available, this group of methods \cite{zhu2024diversified, dhamija2018reducing, hendrycks2017a, du2024does, wang2024learning, liu2020energy, yu2019unsupervised, yang2021semantically, zhang2023mixture} use this data to train the model to improve the network's robustness against OOD data.
    OE \cite{hendrycks2017a} and Energy \cite{liu2020energy, chen2024secure} propose their respective loss functions to be used during the training of the network.
    Recently, \cite{choi2023balanced} improved \cite{liu2020energy} by proposing balanced Energy loss that balances the number of samples of the auxiliary OOD data across classes. 
    Furthermore, \cite{park2022understanding} increases the expected
    Frobenius Jacobian norm difference between ID and OOD. 
    This is in sharp contrast with our intuition as we aim to decrease the norm of the score function in all regions. 
    Another line of work \cite{du2022vos, kong2021opengan, ge2017generative, neal2018open, moon2022difficulty} focuses on generating artificial samples that resemble the real OOD data, using various generative models such as GANs.
    Most recently, OpenMix \cite{Zhu_2023_CVPR} was proposed as a misclassification detection method, teaching the model to reject pseudo-samples generated from the available outlier samples.

    \medskip \noindent 
    \textbf{Outlier Sampling:}
    When large amounts of auxiliary data are available, choosing an informative and diverse set of samples to perform training becomes a priority.
    Many of the state-of-the-art OOD methods use their custom sampling techniques. 
    NTOM \cite{chen2021atom} uses greedy outlier sampling where outliers are selected based on the estimated confidence.
     POEM \cite{ming2022poem} proposes a posterior sampling-based outlier mining framework for OOD detection.
    Most recently DOS \cite{jiang2023diverse} used K-Means \cite{lloyd1982least} to perform clustering on the auxiliary dataset to provide the network with diverse informative OOD samples. 

\section{Preliminaries}

\subsection{Notation}
We consider a classification setup where $\mathcal{X} \in \mathbb{R}^{d}$ denotes the input space and $\mathcal{Y} \in \{1, 2, ..., K\}$ denotes the labels.
We define a neural network $f \colon \mathbb{R}^{d} \rightarrow \mathbb{R}^{K}$ where $f_i(x)$ denotes the $i$-th logit of the neural network to an input $x$. We denote the feature extractor of the neural network as $h$, which is followed by a classification layer. 
We use $f(x) = Wh(x) + b$ for all the models in this paper.
We denote the norm of a vector $x$ as $\|x\|$. 
The log-sum-exp function is defined as $\mathrm{LSE}(x) = \log \sum_{i = 1}^n \exp (x_i)$. For simplicity, for a given score function $S$, we use $S(x)$ as a shorthand for $S(f(x))$.

The In-Distribution (ID), Out-of-Distribution (OOD) and Auxiliary distribution over $\mathcal{X}$ are denoted by  $\mathcal{D}_{\text{in}}$, $\mathcal{D}_{\text{out}}$ and $\mathcal{D}_{\text{aux}}$, respectively. The corresponding datasets are assumed to be sampled i.i.d from their respective distributions. 
We denote the indicator function of an event $e \leq 0$ as $\mathbb{I}_{e \leq 0}$, i.e., if $e\leq 0$  then $\mathbb{I}_{e \leq 0} = 1$, and otherwise, $\mathbb{I}_{e \leq 0} = 0$.
We denote $\mathcal{B}(x, \varepsilon) = \{y | \|y - x\| \leq \varepsilon \}$.
Finally, a function $f$ is said to be locally Lipschitz continuous on $\mathcal{X}$ if there exists a positive constant $L$ such that
$\|f(x) - f(y)\| \leq L \|x - y\| \  \forall x, y \in \mathcal{X}.$
The smallest such constant is called the local Lipschitz constant of $f$.

\subsection{Out-of-Distribution Detection}
It has been observed that when neural networks are faced with samples from a different distribution compared to their training distribution, they produce over-confident errors \cite{hein2019relu}. 
The goal of OOD detection is to differentiate the ID data from the OOD data. 
To achieve this goal, the main approach in the literature is to define a scoring function $S$ and use a threshold $\gamma$ to distinguish different samples.
That is, for a sample $x$, if $S(x) \leq \gamma$, we label it as $\text{ID}$, and label it as $\text{OOD}$ otherwise.

\noindent Most relevant to our setup, \cite{liu2020energy} uses the scoring function $S_{\text{En}}(x) = -\mathrm{LSE}(f(x))$,
% \begin{align}\label{eq:energyScore}
%     ,
% \end{align}
and defines the following loss function to train the network to better differentiate the ID and OOD samples,
\begin{align}\label{eq:energyLoss}
      \mathcal{L}_{S_{\text{En}}} &=
     \mathbb{E}_{(x_{\text{in}}, y_{\text{in}}) \sim \mathcal{D}_{\text{in}}}
    [\mathbb{I}_{S_{\text{En}}(x_{\text{in}}) \geq m_{\text{in}}} (S_{\text{En}}(x_{\text{in}}) - m_{\text{in}}) ^ 2] \\
    \notag&+
    \mathbb{E}_{(x_{\text{aux}}, y_{\text{aux}}) \sim \mathcal{D}_{\text{aux}}}
    [\mathbb{I}_{S_{\text{En}}(x_{\text{aux}}) \leq m_{\text{aux}}}(m_{\text{aux}} - S_{\text{En}}(x_{\text{aux}}))^2],
\end{align}
where $m_{\text{in}}$ and $m_{\text{aux}}$ define two thresholds to filter out points that already have an acceptable level of energy, i.e., if the energy score of an ID (OOD) sample is low (high) enough, it will be excluded from the energy loss.
 
\section{Method}
\begin{figure}[t!]
    \begin{center}
        \includegraphics[width=0.8\columnwidth, , trim={0cm 0cm 3cm 0cm}]{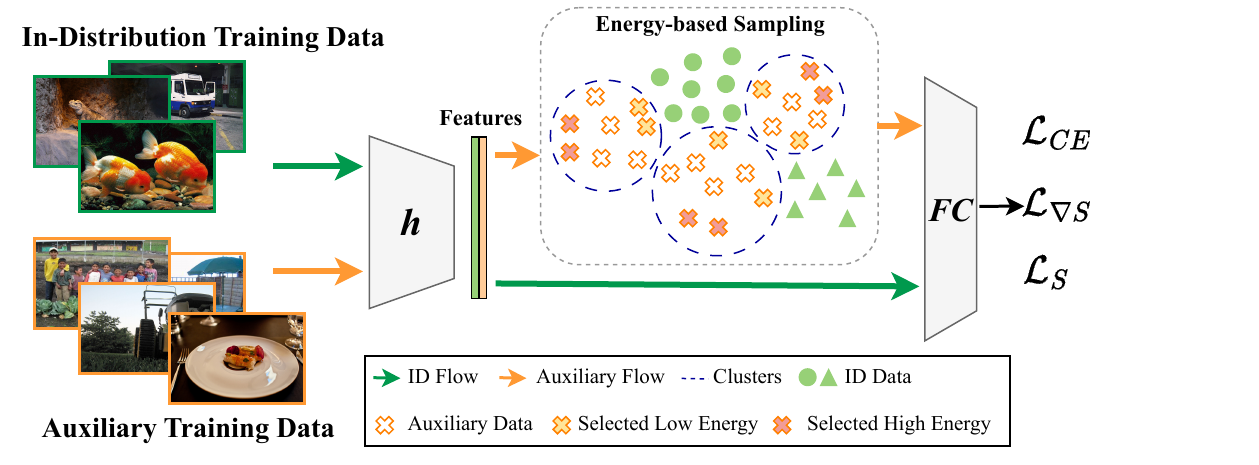}
    \end{center}
    \vskip -15.0pt
    \caption{Overview of \emph{GReg+}. 
    The gradient loss allows the method to obtain more local information from the training data.
    The sampling algorithm uses the normalized features of the OOD samples to perform clustering and choose the sampled data based on the energy score (see \Cref{alg:cluster}) which will be used to calculate the gradient loss to expose the network to samples that can improve the performance of the model.}
    \label{fig:Overview} 
    \vskip -13.0pt
\end{figure}

In this section, we propose our method which consists of regularizing the gradient of the score function by adding a new term to the loss, coupled with a novel energy-based sampling method to allow us to choose more informative samples during training.
% the task of OOD detection.
\Cref{fig:Overview} provides an overview of our method \emph{GReg+} with a simple illustration of how the sampling algorithm utilizes clustering to provide more informative samples.

\subsection{Gradient Regularization}
Many of the current state-of-the-art methods focus only on the value of their scoring function $S$ and aim at minimizing or maximizing it on different regions of the input space \cite{hendrycks2018deep, liu2020energy, choi2023balanced, ming2022poem, jiang2023diverse, Zhu_2023_CVPR}.
We argue that by optimizing only based on the value of the score function, we are not utilizing all of the information that is embedded in the data.
As such, the generalization of the model requires the auxiliary dataset to be a good representation of the actual OOD distribution and its corresponding support in the image space (which is extremely high-dimensional).
In response, we aim to use the local information embedded in the data and propose to do so by leveraging the gradient of the score function $S$. 
In other words, we focus not only on the value of the score function $S$ on the auxiliary samples but also on the local behavior of the score function around these data points.
The rationale behind this approach is that, based on the definition of the problem, the distribution of ID and the OOD tends to be well separated.
Therefore, the area close to an ID sample most likely does not include any OOD samples, and vice versa.
Driven by this insight, we propose to add a regularization term $\mathcal{L}_{\nabla S}$ that promotes the smoothness of the score function around the training samples by penalizing the norm of its gradient.

Suppose $x$ is a training sample. 
Using the first-order Taylor approximation of the score function around $x$, we have
\begin{align}\label{eq:scoreTaylor}
    S(x') \simeq S(x) + \nabla S(x)^\top (x' - x),
\end{align}
for $x'$ sufficiently close to $x$. 
As a result, if $\|\nabla S(x)\|$ is small, then equation \eqref{eq:scoreTaylor} hints that the difference $|S(x')-S(x)|$ would likely remain small, implying stability of the score manifold around the point $x$. 
To this end, we propose the following regularizer.
The specific definition of $\mathcal{L}_{\nabla S}$ is a design choice, however, for the case of the energy loss, which is the main focus of this paper, we use the following formulation.
\begin{align}\label{eq:gradLoss}
    \begin{split}
    \mathcal{L}_{\nabla S_{\text{En}}} 
    &= \mathbb{E}_{(x_{\text{in}}, y_{\text{in}}) \sim \mathcal{D}_{\text{in}}}
    \| \mathbb{I}_{S_{\text{En}}(x_{\text{in}}) \leq m_{\text{in}}}\nabla_{x_{\text{in}}} S_{\text{En}}(x_{\text{in}}) \|\\
    &+
    \mathbb{E}_{(x_{\text{aux}}, y_{\text{aux}}) \sim \mathcal{D}_{\text{aux}}}
    \| \mathbb{I}_{S_{\text{En}}(x_{\text{aux}}) \geq m_{\text{aux}}}\nabla_{x_{\text{aux}}} S_{\text{En}}(x_{\text{aux}}) \|
    \end{split}
\end{align}

\noindent
To elaborate on the intuition behind this choice, consider the following cases:
\begin{enumerate}
    \item If an ID (OOD) sample is correctly detected, we would want the local area around it to be detected the same.
    \item If an ID (OOD) sample is mis-detected as OOD (ID), we do not want its local area to behave the same.
\end{enumerate}
Hence, we select thresholds such that the loss only penalizes the gradient of the correctly detected samples.
This is in sharp contrast to \eqref{eq:energyLoss}, where the aim is to penalize the energy score of the samples that have undesired energy levels.
% Otherwise, we are training the network to have inappropriate values around these inappropriate samples.
Therefore, by designing the gradient loss as \eqref{eq:gradLoss}, we allow the energy loss to train the samples that are not yet well learned and only use the samples whose score values are properly learned.
Having specified $\mathcal{L}_{\nabla S}$, we train our model using the following loss $\mathcal{L} = \mathcal{L}_{\text{CE}} + \lambda_{S}\mathcal{L}_{S} + \lambda_{\nabla S}\mathcal{L}_{\nabla S}, \notag$
where $\lambda_S$ and $\lambda_{\nabla S}$ are regularization constants and $\mathcal{L}_{\text{CE}}$ is the well-known cross-entropy loss.

\subsection{OOD Sampling}\label{sec:oodSampling}
One concern faced by OOD training schemes is that the OOD data may outnumber the ID data by orders of magnitude. 
Having access to such a large auxiliary dataset begs the question of which samples should the network see during the training phase.
This creates two possible scenarios.
Either the training algorithm exposes the model to an equal number of both ID and OOD data points, which means that the model cannot fully utilize the OOD dataset as most of the OOD dataset remains unseen.
Alternatively, the algorithm may provide disproportionately more OOD samples, leading to bias in the model due to the imbalance in the samples \cite{wang2021review}.
These observations hint towards employing a more informative sampling strategy \cite{li2020background, chen2021atom} that utilizes as much of the dataset as possible while avoiding the pitfalls caused by imbalanced datasets. 
Whilst the first option that comes to mind is to perform a greedy sampling based on the score function $S$, it has been observed \cite{jiang2023diverse} that such greedy samples could bias the model to some specific regions of the image space, resulting in a model unable to generalize well.

To address these issues, we present  \Cref{alg:cluster}, which uses clustering (to discourage greedy behavior) alongside energy-based scoring to sample more informative images. Next, we explain the sampling algorithm in more detail.

\smallskip \noindent
\textbf{Clustering:}
Given samples $\{x_i\}_{i = 1}^{n_{\text{OOD}}}$, we first calculate the features $z_i = h(x_i)$ and the scores $s_i = -\text{LSE}(f(x_i))$. 
The next step is to perform clustering in the \emph{feature space}.
To perform clustering, we use K-Means with a fixed number of clusters.
As K-Means assigns clusters based on Euclidean distances, we normalize the features to $\hat{z}_i = \frac{z_i}{\|z_i\|_2}$  to avoid skewed clusters towards samples with disproportionate feature values. 

As for the number of clusters, we use the number of samples in each mini-batch of training, i.e., if in a training iteration the model is presented with $n_{\text{ID}}$ ID samples, we cluster the $n_{\text{OOD}}$ OOD samples into $k = n_{\text{ID}}$ clusters. We study the choice of the number of clusters in the Supplementary Material. 
Finally, given a clustering $\{ C_j \}_{j = 1}^{k}$, we can acquire cluster labels $l_i$ for each data point $x_i, \quad i=1, \cdots, n_{\text{OOD}}$.

\begin{wrapfigure}{r}{0.55\textwidth}
  % \vspace{-45pt} % adjust as necessary to align top
  \begin{minipage}{0.55\textwidth}
    \begin{algorithm}[H]
      \caption{\scriptsize Energy-Based Sampling}
      \label{alg:cluster}
      \textbf{Input:} Auxiliary dataset $\mathcal{X} \sim \mathcal{D}_{\text{aux}}$ with $n_\text{OOD}$ samples,
      number of clusters $k$, 
      feature extractor $h$,
      score function $S$\\
      \textbf{Output:} Selected samples: $\bar{\mathcal{X}} \subset \mathcal{X}$;
      \begin{algorithmic}[1]
          \STATE $z_i \gets h(x_i)$, $s_i \gets S(x_i), i=1, \cdots, n_\text{OOD}.$ 
          \STATE $(\hat{z}_1, \cdots, \hat{z}_{n_\text{OOD}}) \gets h_N(z_1, \cdots, z_{n_\text{OOD}})$
          \STATE Labels $l_i$ $\gets$ K-Means$(\{ \hat{z}_i\}_{i=1}^{n_\text{OOD}}, k)$.
          \FOR{$j=1$ $\cdots$ $k$}
              \STATE $\mathcal{I} \gets \{ i: l_i = j\}$
              \STATE $\bar{\mathcal{X}_j} \gets$ Sample $\{x_i\}_{i \in I}$ based on $\{s_i\}_{i \in I}$
          \ENDFOR
          \STATE return $\bar{\mathcal{X}} \gets \cup_{j=1}^k \bar{\mathcal{X}_j}$
      \end{algorithmic}
    \end{algorithm}
  \end{minipage}
\end{wrapfigure}

\smallskip \noindent
\textbf{Energy-Based Sampling:}
The final step in our sampling algorithm is to choose the \textit{best} samples from each cluster with respect to some criterion. As we use the energy scores $s_i$ to perform OOD detection, we use the same scores for sample selection. 
Given our choice of loss functions in \eqref{eq:energyLoss} and \eqref{eq:gradLoss}, we use the samples with the smallest energy scores $s_i$ of each cluster to provide samples useful for the loss term \eqref{eq:energyLoss}, and we use the samples with the largest energy scores to provide samples that are useful for the loss term \eqref{eq:gradLoss}. Intuitively, $\mathcal{L_{S_\text{En}}}$ aims to increase the energy score of the OOD samples, and so, we need to provide samples whose energy scores are small. On the other hand, $\mathcal{L_{\nabla S_\text{En}}}$ focuses on samples that already have high energy scores to make the model locally behave similarly, and so, we need to also provide samples with high energy scores.

Overall, by clustering, we ensure that the model is exposed to \textit{diverse} regions of the feature space, and by sampling using the scores we ensure that from each region more \textit{informative} samples are used.

    %    \begin{algorithm}[t]
    % \scriptsize
    %   \caption{\scriptsize Energy-Based Sampling}
    %   \label{alg:cluster}
    % \textbf{Input:} Auxiliary dataset $\mathcal{X} \sim \mathcal{D}_{\text{aux}}$ with $n_\text{OOD}$ samples,
    % number of clusters $k$ , 
    % feature extractor $h$,
    % score function $S$\\
    % \textbf{Output:}  Selected samples: $\bar{\mathcal{X}} \subset \mathcal{X}$;
    % \begin{algorithmic}[1]
    %     % \STATE Pass $x \in \mathcal{X}$ through $f$ to obtain the features $z$ and the logits $f(x)$; \\
    %     \STATE $z_i \gets h(x_i)$, $s_i \gets S(x_i), i=1, \cdots, n_\text{OOD}.$ 
    %     % \STATE Normalize the logits $\hat{z} = \frac{z}{\|z\|}$
    %     \STATE $(\hat{z}_1, \cdots, \hat{z}_{n_\text{OOD}}) \gets h_N(z_1, \cdots, z_{n_\text{OOD}})$
    %     % \STATE Run K-Means on the normalized feature space $\hat{z}$ to obtain $k$ clusters $C_j, \: j=1, \cdots k$.\\
    %     \STATE Labels $l_i$ $\gets$ K-Means$(\{ \hat{z}_i\}_{i=1}^{n_\text{OOD}}, k)$.
    %     \FOR{$j=1$ $\cdots$ $k$}
    %         \STATE $\mathcal{I} \gets \{ i: l_i = j\}$
    %         \STATE $\bar{\mathcal{X}_j} \gets$ Sample $\{x_i\}_{i \in I}$ based on $\{s_i\}_{i \in I}$
    %     \ENDFOR \\
    %     % \STATE $$7
    %     \STATE return $\bar{\mathcal{X}} \gets \cup_{j=1}^k \bar{\mathcal{X}_j}$
    % \end{algorithmic}
    % \end{algorithm}

\section{Theoretical Analysis}
\begin{figure}[t]
        \centering
        \includegraphics[width=.8\columnwidth]{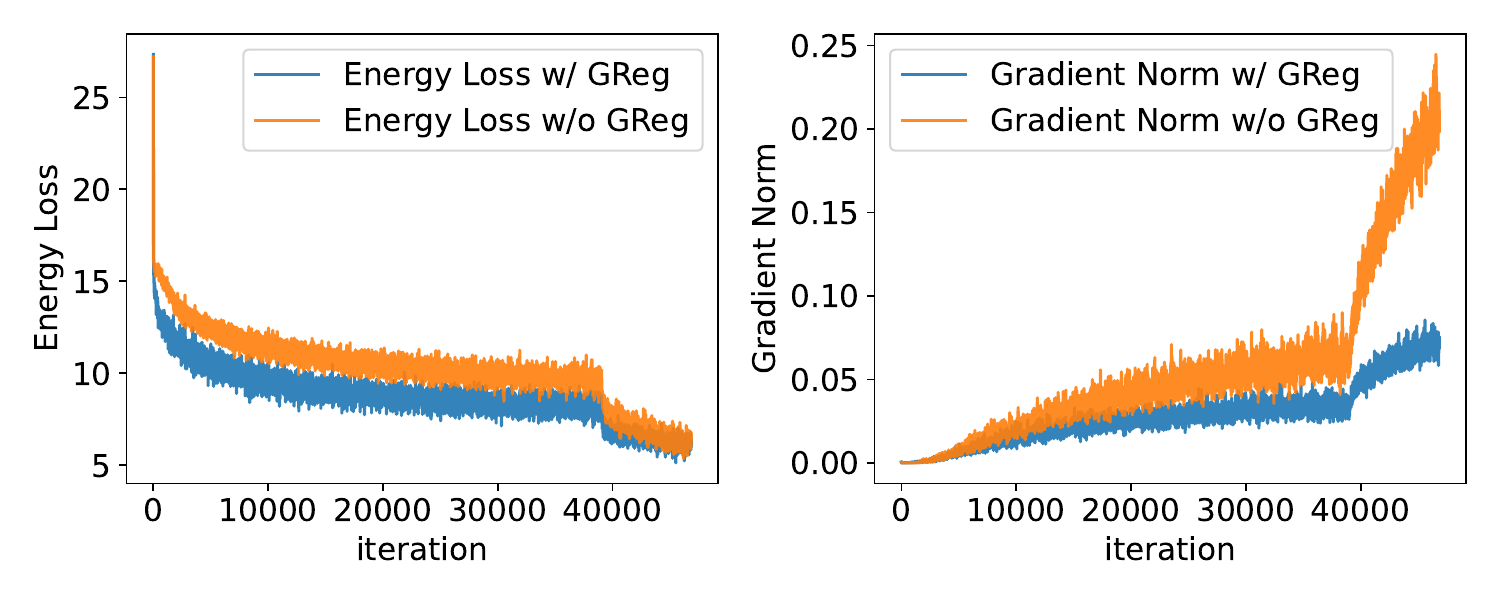}
        \caption{The evolution of the Energy loss and the norm of the gradient with respect to the input of the score function during training with and without gradient regularization. Gradient regularization reduces the slope of the increase of the gradient norm, without negatively affecting the reduction in energy loss. The sudden change in the behavior of the plots towards the final iterations is due to learning rate scheduling.}
        \label{fig:gradientNorm}
    \end{figure}

    In this section, we focus on the theoretical implications of gradient regularization. 
    Traditionally, during the training phase, there is no control over the smoothness of the network and the score function, which can lead to overly sensitive OOD detection. The smoothness of networks has been studied in depth in certified robustness \cite{losch2023certified, srinivas2022efficient, huang2021training, fazlyab2023certified} and has been deemed as a desired or even necessary property for the robustness of neural networks. The desirable properties sought by smoothness consequently follow over to OOD detection; simply because OOD detection can be cast as a simple classification task that the aforementioned literature studies.
    % It can be observed that during the training phase, as the loss of the score function (for example energy loss), decreases, the norm of the gradient of the score function increases,
    Intuitively, the lack of smoothness decreases the network's OOD robustness as the score values can abruptly change with small perturbations in the image space. 
    To support our intuition, consider \Cref{fig:gradientNorm} which portrays the evolution of the energy loss and the norm of its gradient with respect to the input in two cases, with and without \emph{GReg}. 
    Although the energy loss decreases in a similar fashion in both scenarios, the norm of the gradient increases much more rapidly in the absence of gradient regularization. The excess increase in the average norm of the gradient potentially points to a more sensitive score manifold. 
    Note that the general increasing trend of the norm of the gradient in \Cref{fig:gradientNorm}  is not unexpected as the network is initialized randomly.
    
    We further motivate our idea using the concept of penalizing the norm of the gradient \cite{chan2019jacobian} or upper bounds on the norm of the gradient, like the Lipschitz constant \cite{fazlyab2023certified, huang2021training} for achieving robustness.
    In other words, if the gradient of the score function $S$ is bounded around some arbitrary point $x$, there exists an $\varepsilon$-neighborhood in which our prediction of ID/OOD would not change.
    This means that if the point $x$ is detected as ID (OOD), then all the points $x' \in \mathcal{B}(x, \varepsilon)$ are also provably ID (OOD).
    This is desirable for us as it is usually assumed that the ID and OOD data are well separated.
    Formally, suppose $S$ satisfies the following local Lipschitz property on $\mathcal{B}(x, \varepsilon)$ where $x \sim \mathcal{D}_{\text{in}}$
    \begin{align}
        |S(x') - S(x)| \leq L_S\|x' - x\|, \qquad \forall x' \in \mathcal{B}(x, \varepsilon),  \notag
    \end{align}
    where $L_S>0$. 
    Suppose $x \sim \mathcal{D}_{\text{in}}$ is correctly classified by the score function as ID, i.e., $S(x)<\gamma$.
    If we want $x^\prime$ to be also labeled as ID, it is sufficient to satisfy
    $
         S(x) + L_{S} \|x'-x\| \leq \gamma, \notag
    $
    which in turn gives the following certificate
    \begin{equation}\label{eqn:certifiedRadius}
        \|x'-x\| \leq \varepsilon^* = \min \{\varepsilon, \frac{\gamma - S(x)}{L_{S}}\},
    \end{equation}
    
    i.e., all points $x^\prime$ within $\varepsilon^*$ distance to $x$ will also be labeled correctly as ID. A similar argument follows for OOD samples. This bound suggests that we can increase the certified radius by controlling $L_S$ during training, e.g., through penalizing $L_S$.
    There exists a vast literature on Lipschitz estimations for neural networks \cite{fazlyab2019efficient, jordan2020exactly, fazlyab2023certified}.
    However, using these methods on deep models increases the computational load of training. 
    %
    %However, since there exists no Lipschitz calculation method that can produce a reasonable bound for large networks promptly, 
    Instead, we penalize the norm of the gradient, which can be considered as a proxy to the local Lipschitz constant.
    
    Furthermore, for piecewise-linear networks like ReLU and LeakyReLU networks, the local Lipschitz constant is equal to the norm of the gradient. More specifically, for every point $x$\footnote{ \scriptsize We discard the negligible case of points on the boundaries of multiple piecewise-linear spaces.}, there exists an $\hat{\varepsilon} > 0$  such that the network remains linear in this region, i.e., 
    \[
    f(x') = Ax' + b, \qquad \forall x' \in \mathcal{B}(x, \hat{\varepsilon}).
    \]
    In such a region, the local Lipschitz constant is equal to the norm of the gradient of the points \cite{bhowmick2021lipbab}, i.e., $L_f = \| A \|$. 
    This further motivates directly penalizing the norm of the gradient of $S$.
    Thus, for such networks, minimizing $\| \nabla S(x) \|$ decreases the local Lipschitz constant in the neighborhood of $x$, which in turn could increase the maximum certified perturbation radius in \eqref{eqn:certifiedRadius}. Note that our argument for using the norm of the gradient in practice does not require the knowledge of $\hat{\varepsilon}$. We just use the fact that such an $\hat{\varepsilon}$ exists and that in the corresponding region, the network will be linear, and as a result, minimizing the norm of the gradient is equivalent to minimizing the local Lipschitz constant for this local neighborhood.
    
    \medskip \noindent
    \textbf{Post-hoc Methods and Gradient Regularization:}
        So far, we have 
        % discussed the idea of gradient regularization and 
        provided a certified robustness radius against OOD misdetection using the Lipschitz constant of the network.
        This radius \eqref{eqn:certifiedRadius} is proportional to the inverse of the Lipschitz constant of the score function, and by regularizing the gradient we aim to reduce the Lipschitz constant to increase this radius.
        Here, we note that many of the state-of-the-art OOD detection methods can also be analyzed in this framework. For example,  DICE \cite{sun2022dice} and LINe \cite{ahn2023line} use activation pruning and sparsification to remove noisy neurons. 
        It has been shown \cite{huang2021training} that for a given matrix, removing a row or column of it results in a reduction of the Lipschitz constant. 
        As activation pruning is essentially a form of masking the weights of the network, methods utilizing activation pruning reduce the upper bound on the Lipschitz constant, acquiring a network with a smaller Lipschitz upper bound.

\section{Experiments}
We perform extensive OOD detection experiments on CIFAR  \cite{krizhevsky2009learning} and ImageNet \cite{deng2009imagenet} benchmarks. 
In our comparisons, we use a wide range of CNN-based architectures.
We also provide ablation analysis to demonstrate the effect of each component within our framework.

\subsection{Setup}
\textbf{Datasets.} 
For CIFAR experiments, we use CIFAR-10 and CIFAR-100 as ID datasets. We evaluate the models on six different OOD datasets including Textures \cite{cimpoi2014describing}, SVHN \cite{netzer2011reading}, Places365 \cite{zhou2017places}, LSUN-cropped, LSUN-resized \cite{yu2015lsun}, and iSUN \cite{xu2015turkergaze}. 
Following \cite{sun2022dice}, at test time, we use 10000 randomly selected samples from each of these 6 datasets.
Following \cite{Zhu_2023_CVPR}, instead of the recently retracted TinyImages, we use the 300K RandomImages \cite{hendrycks2018deep} as the auxiliary outlier dataset. 

For the ImageNet experiments, following \cite{jiang2023diverse}, we consider a set of 10 \emph{random} classes from ImageNet-1K as the ID dataset and the rest of the dataset as the auxiliary outlier dataset. 
For evaluation, we use 10000 randomly selected samples from Textures, Places, SUN \cite{xiao2010sun}, and iNaturalist \cite{van2018inaturalist} as OOD datasets.

\begin{table*}[t!]
	\caption{\textbf{Comparison on the CIFAR-10 benchmark.} Mean AUROC and FPR95 percentages on various benchmarks. The experimental results are reported over three trials. The best mean results are bolded and the runner-up is underlined. AUROC and FPR95 are percentages.}
	\vskip -0.07in
	\label{tab:cifar10}
	\setlength\tabcolsep{6pt}
	\centering
	\renewcommand{\arraystretch}{1}
	\scalebox{0.64}{
		\begin{tabular}{llccccccccc}
			\toprule
			\multirow{2}{*}{\textbf{Dataset}} &\multirow{2}{*}{\textbf{Method}} & \multicolumn{2}{c}{\textbf{ResNet}} && \multicolumn{2}{c}{\textbf{WRN}} && \multicolumn{2}{c}{\textbf{DenseNet}} \\
			\cmidrule(lr){3-5}  \cmidrule(lr){6-7} \cmidrule(lr){9-11}
			& & \textbf{FPR95} $\downarrow$ & \textbf{AUROC} $\uparrow$  &&\textbf{FPR95} $\downarrow$ & \textbf{AUROC} $\uparrow$ &&\textbf{FPR95} $\downarrow$ & \textbf{AUROC} $\uparrow$ \\
			\midrule
   
			\multirow{12}{*}{CIFAR-10} 
			&MSP~{\scriptsize\textcolor{darkgray}{[ICLR17]}} \cite{hendrycks2017a}  
                    & 58.04 & 90.49 && 51.52 & 90.8 && 52.56 & 91.7  \\
			&ODIN~{\scriptsize\textcolor{darkgray}{[ICLR18]}} \cite{liang2017enhancing}  
                    &34.46 & 91.59 && 35.23 & 89.82 && 24.88 & 94.42  \\
                &Energy Score~{\scriptsize\textcolor{darkgray}{[Neurips20]}} \cite{liu2020energy}  
                    & 39.32 & 92.24 && 33.4 & 91.76 && 28.99 & 94.09  \\
			&ReAct~{\scriptsize\textcolor{darkgray}{[NeurIPS21]}} \cite{sun2021react}  
                    & 39.44 & 92.30 && 37.54 & 91.49 && 25.83 & 95.27   \\
                &Dice~{\scriptsize\textcolor{darkgray}{[ECCV22]}} \cite{sun2022dice}  
                     &42.40 & 91.25 && 34.12 & 91.74 && 26.37 &	94.61\\
                &LINe~{\scriptsize\textcolor{darkgray}{[CVPR23]}} \cite{ahn2023line}  
                    &45.25 & 90.59 && 36.27 & 90.2 && 14.84 & 96.95  \\
                \cmidrule(lr){2-11}
			&OE~{\scriptsize\textcolor{darkgray}{[ICLR18]}} \cite{hendrycks2018deep} 
                    &19.92 & 95.71 && 21.12 & 95.55 && 21.76 & 95.8   \\
			&Energy Loss~{\scriptsize\textcolor{darkgray}{[NeurIPS20]}} \cite{liu2020energy} 
                    &\underline{11.14} & \underline{97.53} && \underline{13.11} & \underline{97.14}  && \underline{11.26} & \underline{97.43}  \\
            &OpenMix~{\scriptsize\textcolor{darkgray}{[CVPR23]}} \cite{Zhu_2023_CVPR} 
                    &22.24 & 96.26 && 21.92 & 96 && 22.86 & 95.65   \\
                \cmidrule(lr){2-11}
                &\emph{GReg} 
                    &\textbf{7.9} & \textbf{97.95} && \textbf{7.95} & \textbf{98.1} && \textbf{7.93} & \textbf{98.12}   \\
			\midrule
		\end{tabular}
	}
	%\vskip -0.1 in
\end{table*}

\begin{table*}[t]
	\caption{\textbf{Comparison on the CIFAR-100 benchmark.} Mean AUROC and FPR95 percentages on various benchmarks. The experimental results are reported over three trials. The best mean results are bolded and the runner-up is underlined. }
	\vskip -0.07in
	\label{tab:cifar100}
	\setlength\tabcolsep{6pt}
	\centering
	\renewcommand{\arraystretch}{1}
	\scalebox{0.63}{
		\begin{tabular}{llccccccccc}
			\toprule
			\multirow{2}{*}{\textbf{Dataset}} &\multirow{2}{*}{\textbf{Method}} & \multicolumn{2}{c}{\textbf{ResNet}} && \multicolumn{2}{c}{\textbf{WRN}} && \multicolumn{2}{c}{\textbf{DenseNet}} \\
			\cmidrule(lr){3-5}  \cmidrule(lr){6-7} \cmidrule(lr){9-11}
			& & \textbf{FPR95} $\downarrow$ & \textbf{AUROC} $\uparrow$  &&\textbf{FPR95} $\downarrow$ & \textbf{AUROC} $\uparrow$ &&\textbf{FPR95} $\downarrow$ & \textbf{AUROC} $\uparrow$ \\
			\midrule
            \multirow{14}{*}{CIFAR-100} 
			&MSP~{\scriptsize\textcolor{darkgray}{[ICLR17]}} \cite{hendrycks2017a}  
                    &74.66	& 80.10 && 80.39 & 75.37 && 85.87 & 68.78   \\
			&ODIN~{\scriptsize\textcolor{darkgray}{[ICLR18]}} \cite{liang2017enhancing}  
                    &64.00 & 84.33 && 67.95 & 79.86 && 69.97 & 81.55   \\
                &Energy Score~{\scriptsize\textcolor{darkgray}{[Neurips20]}} \cite{liu2020energy}  
                    &68.46 & 84.15 && 74.15 & 79.26 && 78.91 & 76.12   \\
			&ReAct~{\scriptsize\textcolor{darkgray}{[NeurIPS21]}} \cite{sun2021react}  
                    &59.38 & 87.59 && 70.65 & 80.2 && 76.27 & 80.1   \\
                &Dice~{\scriptsize\textcolor{darkgray}{[ECCV22]}} \cite{sun2022dice}  
                    &77.29 & 80.35 && 72.32 & 79.62 && 69.80 & 80.50   \\
                &LINe~{\scriptsize\textcolor{darkgray}{[CVPR23]}} \cite{ahn2023line}  
                    &72.09 & 82.93 && 67.5 & 80.9 && 35.16 & 88.72   \\
                \cmidrule(lr){2-11}
			&OE~{\scriptsize\textcolor{darkgray}{[ICLR18]}} \cite{hendrycks2018deep} 
                    &77.32 & 63.89 && 74.22 & 65.70 && 60.52 & 84.21   \\
			&Energy Loss~{\scriptsize\textcolor{darkgray}{[NeurIPS20]}} \cite{liu2020energy} 
                    &62.77 & 84.05 && 65.62 & 80.18  && 64.37 & 83.86   \\
			&POEM~{\scriptsize\textcolor{darkgray}{[ICML22]}} \cite{ming2022poem} 
                    &61.11 & 82.43 && 62.58 & 79.74 && 59.33 & 79.81   \\
                &OpenMix~{\scriptsize\textcolor{darkgray}{[CVPR23]}} \cite{Zhu_2023_CVPR} 
                    & 59.64 & 88.09 && 73.12 & 79.23 && 67.29 & 84.47   \\
                &DOS~{\scriptsize\textcolor{darkgray}{[ICLR24]}} \cite{jiang2023diverse} 
                    & \underline{54.62} & \underline{88.30} &&\textbf{45.26} & \underline{90.76} && \underline{34.92} & \underline{93.22}  \\
                \cmidrule(lr){2-11}
                &\emph{GReg} 
                    & 59.6 & 82.92 && 58.26 & 86.56 && 56.29 & 87.35   \\
                &\emph{GReg+}
                    &\textbf{50.78} & \textbf{88.75} && \underline{48.12} & \textbf{91.02} &&\textbf{30.55} & \textbf{93.38}   \\
			\midrule
		\end{tabular}
	}
	%\vskip -0.1 in
\end{table*}

\begin{table*}[t!]
\centering
\caption{\textbf{Comparison on the ImageNet benchmark.} The experimental results are reported over three trials. The best mean results are bolded and the runner-up is underlined. AUROC and FPR95 are percentages.
} 
\label{tab:imagenet}
\scalebox{0.64}{
\begin{tabular}{lcccccccccc}
    \toprule
  \multirow{3}{*}{\textbf{Method}}  & \multicolumn{8}{c}{\textbf{OOD Datasets}} & \multicolumn{2}{c}{\multirow{2}{*}{\textbf{Average}}} \\ \cline{2-9}
 \multicolumn{1}{c}{} & \multicolumn{2}{c}{\textbf{iNaturalist}} & \multicolumn{2}{c}{\textbf{SUN}} & \multicolumn{2}{c}{\textbf{Places}} & \multicolumn{2}{c}{\textbf{Textures}} & & \\
 \multicolumn{1}{c}{} & FPR95 $\downarrow$  & AUROC $\uparrow$ & FPR95 $\downarrow$ & AUROC $\uparrow$ & FPR95 $\downarrow$ & AUROC $\uparrow$ & FPR95 $\downarrow$ & AUROC $\uparrow$ & FPR95 $\downarrow$ & AUROC $\uparrow$ \\
 \hline
MSP~{\scriptsize\textcolor{darkgray}{[ICLR17]}} \cite{hendrycks2017a} & 62.6 & 87.09 & 64.3 &	86.64 &	62.3 &	86.72 &	78.3 &	75.03 &	66.87 &	83.87 \\
ODIN~{\scriptsize\textcolor{darkgray}{[ICLR18]}} \cite{liang2017enhancing} & 48.5 &	92.07 &	53.9 &	 90.77 &	49.3 &	91.06 &	69.6 &	80.77 &	55.32 &	88.66\\
Energy Score~{\scriptsize\textcolor{darkgray}{[Neurips20]}}\cite{liu2020energy} & 54 &	90.77 &	52.5 &	90.7 &	46.6 &	91.37 &	73.1 &	77.54 &	56.55 &	87.59\\
ReAct~{\scriptsize\textcolor{darkgray}{[NeurIPS21]}} \cite{sun2021react} & 51.88 &	91.52 &	52.48 &	90.81 & 49.56 & 90.92 & 64.61 & 83.77 & 54.63 & 89.25\\
DICE~{\scriptsize\textcolor{darkgray}{[ECCV22]}}\cite{sun2022dice} & 32.86 & 93.6 &	41.35 &	92.9 &	45.82 &	91.58 &	65.2 &	78.61 &	46.3 & 89.17\\
LINe~{\scriptsize\textcolor{darkgray}{[CVPR23]}}\cite{ahn2023line} &  29.47 & 93.77 & 37.29 &	92.85 & 38.85 & 92.16 & 52.32 & 86.38 & \underline{39.48} & 91.29\\
\midrule
OE~{\scriptsize\textcolor{darkgray}{[ICLR18]}}\cite{hendrycks2018deep} & 34.56 & 92.75 & 54.63 & 89.75 & 54.86 &  89.05 & 76.2 & 75.55 &	55.06 &	86.78 \\
Energy Loss~{\scriptsize\textcolor{darkgray}{[Neurips20]}}\cite{liu2020energy} & 18.44 & 95.91 & 50.19 & 90.37 & 49.32 &	90.47 &	62.96 &	77.78 &	45.23 &	88.63\\
DOS~{\scriptsize\textcolor{darkgray}{[ICLR24]}}\cite{jiang2023diverse} & 61.63 & 88.68 &	42.69 &	93.23 &	45.49 &	93.37 &	47.45 &	92.63 & 49.31 &	\underline{91.97}\\
\midrule
\textit{GReg} & 29.54 &	94.94 &	45.86 &	92.53 &	45.35 &	92.06 &	68.29 &	80.98 &	47.26 &	90.13\\
\textit{GReg+} & 24.11	&95&	33.98&	92.31& 32.89&	92.7& 49.32&	88.22& \textbf{35.08}&	\textbf{92.06}\\
\bottomrule
\end{tabular}
}
\end{table*}

\smallskip \noindent
\textbf{Experimental Setup:}
For the experiments of \emph{GReg} on CIFAR, we use a pre-trained model of the corresponding architecture and finetune for 20 epochs using SGD with cosine annealing. 
For the experiments of \emph{GReg+} on CIFAR, we train a model from scratch for 50 epochs with a learning rate of $0.1$ and then further train for another 10 epochs with a learning rate of $0.01$, using the same optimizer.
Following \cite{liu2020energy}, we set $\lambda_S = 0.1 $ and set $\lambda_{\nabla S} = 1$ .
\\
For the ImageNet experiment, we only perform finetuning on a pre-trained DenseNet-121. See the Supplementary Material for more details.\\
For the CIFAR experiments, we evaluate our results on ResNet-18 \cite{he2016deep}, WRN-40 \cite{zagoruyko2016wide}, and DenseNet-101 \cite{huang2017densely} to span a wide range of architectures and parameter numbers.

\smallskip \noindent
\textbf{Sampling Details:} 
Following the explanations in \Cref{sec:oodSampling}, we cluster the auxiliary samples into $n_{\text{ID}}$ clusters and choose the lowest and highest scoring samples of each cluster, i.e., in each iteration, we present $2n_{\text{ID}}$ auxiliary samples for the training of that iteration.
For the CIFAR experiments, we use the whole auxiliary dataset in each epoch for clustering.
In other words, at each epoch, the 300K images are split into mini-batches, and clustering is performed on each mini-batch at each iteration so that no auxiliary sample is missed.
% This means the clustering is performed on each mini-batch and not the full auxiliary dataset, as doing so would be computationally prohibitive.
For the ImageNet experiment, in each epoch, we sample 600K randomly selected images from the remaining (990) classes as the auxiliary dataset and perform the clustering on each mini-batch.

\smallskip\noindent
\textbf{Evaluation Metrics:}
We report the following metrics: 1) \emph{FPR95}: the false positive rate of the OOD samples when the true positive rate of ID samples is at 95\%.
2) \emph{AUROC}: the area under the receiver operating characteristic curve.

\smallskip\noindent
\textbf{Comparison Methods:}
For comparison with post-hoc OOD methods, we compare to MSP \cite{hendrycks2017a}, ODIN \cite{hendrycks2018deep}, Energy (score) \cite{liu2020energy}, ReAct \cite{sun2021react}, DICE \cite{sun2022dice} and LINe \cite{ahn2023line}.
For the methods that use auxiliary data, we compare with 
OE \cite{hendrycks2018deep}, Energy (loss) \cite{liu2020energy}, POEM \cite{ming2022poem}, OpenMix \cite{Zhu_2023_CVPR} and DOS \cite{jiang2023diverse}.
MSP uses the maximum softmax probability to detect the OOD samples. ODIN improves upon MSP by introducing noise and temperature scaling. Energy proposes a score function for OOD detection. 
ReAct uses activation clipping. DICE and LINe build up on that idea by also detecting the important neurons and activation to improve OOD detection.
OE and Energy train based on their respective loss functions.
POEM uses posterior sampling and DOS proposes diverse sampling to improve OOD detection.
OpenMix uses pseudo-samples to improve misclassification detection. 

\begin{table}[t!]
	\caption{\textbf{Ablation Study of \emph{GReg}.} 
            AUROC and FPR95 percentages on CIFAR benchmarks averaged over all OOD datasets and three runs on DenseNet.}
	\vskip -0.07in
	\label{tab:ablationGrad}
	\setlength\tabcolsep{6pt}
	\centering
	\renewcommand{\arraystretch}{1}
	\scalebox{0.7}{
		\begin{tabular}{llccccccccc}
			\toprule
			\multirow{2}{*}{\textbf{Method}} & \multicolumn{2}{c}{\textbf{CIFAR-10}} && \multicolumn{2}{c}{\textbf{CIFAR-100}}\\
			\cmidrule(lr){2-4}  \cmidrule(lr){5-6}
			 & \textbf{FPR95} $\downarrow$ & \textbf{AUROC} $\uparrow$  &&\textbf{FPR95} $\downarrow$ & \textbf{AUROC} $\uparrow$  \\
			\midrule			
			OE
                    & 21.76 & 95.8 && \textbf{60.52} & 84.21   \\
                OE + Grad
                    & \textbf{18.79} & \textbf{96.32} && 62.54 & \textbf{84.71}   \\
                \cmidrule(lr){2-11}
			Energy 
                    & 11.26 & 97.43 && 64.37 & 83.86   \\
                Energy + Grad
                    & \textbf{7.93} & \textbf{98.12} && \textbf{56.29} & \textbf{87.35}   \\
                \cmidrule(lr){2-11}
			POEM 
                    & 29.15 & 94.18 && 59.33 & 79.81   \\
                POEM + Grad
                    & \textbf{23.87} &\textbf{95.41} && \textbf{47.94} & \textbf{86.63}  \\
			\bottomrule
		\end{tabular}
	} 
\end{table}

\subsection{Results}
\textbf{Evaluation on CIFAR-10:} 
We first evaluate the performance of our method on the CIFAR-10 dataset. 
Following the literature \cite{ahn2023line}, Table \ref{tab:cifar10} shows the average results over the six commonly used OOD datasets. The detailed results on individual OOD datasets can be found in the Supplementary Material. 
As the results on CIFAR-10 are saturated, in \Cref{tab:cifar10} we focus on the effectiveness of gradient regularization and report the rest of the methods in the Supplementary Material.
It can be seen that \emph{GReg} outperforms all the methods in both of the reported metrics. 
Specifically, \emph{GReg} decreases the FPR95 of \cite{liu2020energy} by 3.9\%. 

\smallskip \noindent
\textbf{Evaluation on CIFAR-100:} 
For the results of this part, we use the same experimental setup and only change the ID dataset to CIFAR-100.
The average results of this experiment set are presented in \Cref{tab:cifar100} and the individual results can be found in the Supplementary Material. 
In this setup, \emph{GReg} provides competitive results on the ResNet and WRN architectures with state-of-the-art methods that do not employ sampling.
Moreover, coupled with sampling, \emph{GReg+} outperforms all current state-of-the-art methods on almost all metrics of the different architectures. On average, \emph{GReg+} outperforms the FPR95 metric of DOS \cite{jiang2023diverse} by 1.8\%, which showcases the effectiveness of energy-based sampling.
Results on each OOD dataset can be found in the Supplementary Material.

\smallskip \noindent
\textbf{Evaluation on ImageNet:}
For this experiment, we randomly select 10 classes of Imagenet-1K as ID and use the rest as the auxiliary data. 
The results of this large-scale experiment are presented in Table \ref{tab:imagenet}. We can see that in terms of FPR95, \emph{GReg+} outperforms DOS and LINe by large margins of $14.23\%$ and $3.68\%$, respectively, while achieving state-of-the-art AUROC performance. 
The energy-based sampling approach enables our method to acquire more informative samples compared to the sampling method in \cite{jiang2023diverse}, resulting in improved FPR95.

\subsection{Ablation Study}
In this section, we perform extensive experiments to show the effectiveness of components within our framework, individually. 
That is, we show the effect of regularizing the gradient on other methods and then compare our proposed sampling method with the state-of-the-art. 
We also study the effect of our OOD schemes on the ID accuracy of the models.

\smallskip \noindent
\textbf{Gradient Regularization:}
Due to the generality of the idea of gradient regularization, our method can be used in conjunction with many existing methods.
% These methods include but are not limited to, Energy \cite{liu2020energy}, OE \cite{hendrycks2018deep}, POEM \cite{ming2022poem}, etc.
To leverage the gradient regularization idea, one needs to define a suitable loss function based on the norm of the gradient of the score, to promote a smoother score manifold.
% Once the gradient loss is designed, adding this term to the total loss function promotes a smoother score manifold learned by the neural network.
    % 
To show the effectiveness of \emph{GReg} on different OOD training schemes, we choose three well-known OOD detection methods, Energy, OE, and POEM, and train a DenseNet with and without gradient regularization on both CIFAR benchmarks. 
\Cref{tab:ablationGrad} shows the overall improvement of all these methods with gradient regularization, in both the FPR95 and AUROC, supporting our claim that coupling \emph{GReg} with other methods could improve their performance. 

\smallskip \noindent
\textbf{Sampling Methods:}
To study the effect of various sampling methods, we compare POEM and DOS to our method with and without gradient regularization.
We denote \emph{GReg+} without gradient regularization as Energy-only-clustering.
\Cref{tab:ablationSampling} compares different methods that utilize some form of sampling and shows the impact of energy-based clustering.
On top of that, the comparison between Energy-only-clustering and \emph{GReg+} further encourages gradient regularization.

\smallskip \noindent
\textbf{ID Accuracy:} 
To study the effect of gradient regularization on accuracy, we compare the ID accuracy of our methods to the most relevant state-of-the-art on WRN and DenseNet architectures.
% to ensure that our method does not have a large negative impact on the ID accuracy.
\Cref{tab:idAccuracy} shows that gradient regularization (\emph{GReg}) improves the ID accuracy of energy on both architectures and \emph{GReg+} also has superior ID accuracy in comparison with DOS.
\begin{table}[!t]
    \centering
    \begin{minipage}[t]{0.45\linewidth}
    \caption{\textbf{Ablation Study on Sampling.} 
                    Comparison among sampling methods on CIFAR-100 on DenseNet averaged over six OOD datasets.}
	\vskip -0.07in
	\label{tab:ablationSampling}
	\setlength\tabcolsep{6pt}
    \scalebox{0.7}{
		\begin{tabular}{llccccccccc}
			\toprule
			\multirow{2}{*}{\textbf{Method}} &\multicolumn{2}{c}{\textbf{CIFAR-100}}\\
			\cmidrule(lr){2-4}
			 & \textbf{FPR95} $\downarrow$ & \textbf{AUROC} $\uparrow$ \\
			\midrule
		   POEM &  59.33 &  79.81 \\
			 DOS &  34.92 & 93.22  \\
			 Energy only clustering & 31.77	& 93.20   \\
			 \emph{GReg+} &  \textbf{30.55} & \textbf{93.38}   \\
			\bottomrule
		\end{tabular}
	} 
    \end{minipage}
    \hspace{3mm}
    \begin{minipage}[t]{0.45\linewidth}
        \caption{\textbf{Ablation Study on ID Accuracy.} 
                    Accuracy comparison on CIFAR-10 on two architectures.}
	\vskip -0.07in
	\label{tab:idAccuracy}
	\setlength\tabcolsep{6pt}
	\centering
	\renewcommand{\arraystretch}{1}
    \scalebox{0.7}{
		\begin{tabular}{llccccccccc}
			\toprule
			\multirow{2}{*}{\textbf{Method}} &\multicolumn{2}{c}{\textbf{CIFAR-10}}\\
			\cmidrule(lr){2-4}
			 & \textbf{WRN}  & \textbf{DenseNet} \\
			\midrule
		   Clean &
                            94.84 &  94.03 \\
		   Energy loss &
                            88.15 &  91.82 \\
		   \emph{GReg}  &
                            89.55 &  92.14 \\
		   DOS &
                            78.90 &  87.06 \\
			 \emph{GReg+} &
                            83.06 & 89.89   \\
			\bottomrule
		\end{tabular}
	} 
    \end{minipage}
\end{table}

\section{Conclusions}
In this paper, we presented the idea of gradient regularization of the score function in OOD detection. 
We also developed an energy-based sampling algorithm to improve sampling quality when the auxiliary dataset is large. 
We demonstrate that \emph{GReg} exhibits superior performance compared to methods that do not require clustering, and \emph{GReg+} outperforms all state-of-the-art methods. 
Furthermore, our experiments show that regularizing the gradient could result in robust models capable of extracting local information from the ID and OOD data.
We also provide detailed analytical analysis and ablation studies to support our method. 

\section* {Acknowledgements}
Research was sponsored by the Army Research Laboratory and was accomplished under Cooperative Agreement Number W911NF-23-2-0008. The views and conclusions contained in this document are those of the authors and should not be interpreted as representing the official policies, either expressed or implied, of the Army Research Laboratory or the U.S. Government. The U.S. Government is authorized to reproduce and distribute reprints for Government purposes notwithstanding any copyright notation herein.

\bibliographystyle{splncs04}
\bibliography{main}

\begin{thebibliography}{10}
\providecommand{\url}[1]{\texttt{#1}}
\providecommand{\urlprefix}{URL }
\providecommand{\doi}[1]{https://doi.org/#1}

\bibitem{ahn2023line}
Ahn, Y.H., Park, G.M., Kim, S.T.: Line: Out-of-distribution detection by leveraging important neurons. arXiv preprint arXiv:2303.13995  (2023)

\bibitem{Bafghi_2023_CVPR}
Bafghi, R.A., Gurari, D.: A new dataset based on images taken by blind people for testing the robustness of image classification models trained for imagenet categories. In: Proceedings of the IEEE/CVF Conference on Computer Vision and Pattern Recognition (CVPR). pp. 16261--16270 (June 2023)

\bibitem{bhowmick2021lipbab}
Bhowmick, A., D’Souza, M., Raghavan, G.S.: Lipbab: Computing exact lipschitz constant of relu networks. In: Artificial Neural Networks and Machine Learning--ICANN 2021: 30th International Conference on Artificial Neural Networks, Bratislava, Slovakia, September 14--17, 2021, Proceedings, Part IV 30. pp. 151--162. Springer (2021)

\bibitem{cao2024envisioning}
Cao, C., Zhong, Z., Zhou, Z., Liu, Y., Liu, T., Han, B.: Envisioning outlier exposure by large language models for out-of-distribution detection. arXiv preprint arXiv:2406.00806  (2024)

\bibitem{chan2019jacobian}
Chan, A., Tay, Y., Ong, Y.S., Fu, J.: Jacobian adversarially regularized networks for robustness. arXiv preprint arXiv:1912.10185  (2019)

\bibitem{chen2020learning}
Chen, G., Qiao, L., Shi, Y., Peng, P., Li, J., Huang, T., Pu, S., Tian, Y.: Learning open set network with discriminative reciprocal points. In: Computer Vision--ECCV 2020: 16th European Conference, Glasgow, UK, August 23--28, 2020, Proceedings, Part III 16. pp. 507--522. Springer (2020)

\bibitem{chen2021atom}
Chen, J., Li, Y., Wu, X., Liang, Y., Jha, S.: Atom: Robustifying out-of-distribution detection using outlier mining. In: Machine Learning and Knowledge Discovery in Databases. Research Track: European Conference, ECML PKDD 2021, Bilbao, Spain, September 13--17, 2021, Proceedings, Part III 21. pp. 430--445. Springer (2021)

\bibitem{chen2024secure}
Chen, S., Huang, L.K., Schwarz, J.R., Du, Y., Wei, Y.: Secure out-of-distribution task generalization with energy-based models. Advances in Neural Information Processing Systems  \textbf{36} (2024)

\bibitem{choi2023balanced}
Choi, H., Jeong, H., Choi, J.Y.: Balanced energy regularization loss for out-of-distribution detection. In: Proceedings of the IEEE/CVF Conference on Computer Vision and Pattern Recognition. pp. 15691--15700 (2023)

\bibitem{choi2023projection}
Choi, S., Lee, H., Lee, H., Lee, M.: Projection regret: Reducing background bias for novelty detection via diffusion models. Advances in Neural Information Processing Systems  \textbf{36},  19230--19245 (2023)

\bibitem{cimpoi2014describing}
Cimpoi, M., Maji, S., Kokkinos, I., Mohamed, S., Vedaldi, A.: Describing textures in the wild. In: Proceedings of the IEEE conference on computer vision and pattern recognition. pp. 3606--3613 (2014)

\bibitem{deng2009imagenet}
Deng, J., Dong, W., Socher, R., Li, L.J., Li, K., Fei-Fei, L.: Imagenet: A large-scale hierarchical image database. In: 2009 IEEE conference on computer vision and pattern recognition. pp. 248--255. Ieee (2009)

\bibitem{dhamija2018reducing}
Dhamija, A.R., G{\"u}nther, M., Boult, T.: Reducing network agnostophobia. Advances in Neural Information Processing Systems  \textbf{31} (2018)

\bibitem{du2024does}
Du, X., Fang, Z., Diakonikolas, I., Li, Y.: How does unlabeled data provably help out-of-distribution detection? arXiv preprint arXiv:2402.03502  (2024)

\bibitem{du2022vos}
Du, X., Wang, Z., Cai, M., Li, Y.: Vos: Learning what you don't know by virtual outlier synthesis. arXiv preprint arXiv:2202.01197  (2022)

\bibitem{Fan_2024_CVPR}
Fan, K., Liu, T., Qiu, X., Wang, Y., Huai, L., Shangguan, Z., Gou, S., Liu, F., Fu, Y., Fu, Y., Jiang, X.: Test-time linear out-of-distribution detection. In: Proceedings of the IEEE/CVF Conference on Computer Vision and Pattern Recognition (CVPR). pp. 23752--23761 (June 2024)

\bibitem{fazlyab2023certified}
Fazlyab, M., Entesari, T., Roy, A., Chellappa, R.: Certified robustness via dynamic margin maximization and improved lipschitz regularization (2023)

\bibitem{fazlyab2019efficient}
Fazlyab, M., Robey, A., Hassani, H., Morari, M., Pappas, G.: Efficient and accurate estimation of lipschitz constants for deep neural networks. In: Advances in Neural Information Processing Systems. pp. 11427--11438 (2019)

\bibitem{ge2017generative}
Ge, Z., Demyanov, S., Chen, Z., Garnavi, R.: Generative openmax for multi-class open set classification. arXiv preprint arXiv:1707.07418  (2017)

\bibitem{he2016deep}
He, K., Zhang, X., Ren, S., Sun, J.: Deep residual learning for image recognition. In: Proceedings of the IEEE conference on computer vision and pattern recognition. pp. 770--778 (2016)

\bibitem{hein2019relu}
Hein, M., Andriushchenko, M., Bitterwolf, J.: Why relu networks yield high-confidence predictions far away from the training data and how to mitigate the problem. In: Proceedings of the IEEE/CVF Conference on Computer Vision and Pattern Recognition. pp. 41--50 (2019)

\bibitem{hendrycks2017a}
Hendrycks, D., Gimpel, K.: A baseline for detecting misclassified and out-of-distribution examples in neural networks. In: International Conference on Learning Representations (2017), \url{https://openreview.net/forum?id=Hkg4TI9xl}

\bibitem{hendrycks2018deep}
Hendrycks, D., Mazeika, M., Dietterich, T.: Deep anomaly detection with outlier exposure. In: International Conference on Learning Representations (2019), \url{https://openreview.net/forum?id=HyxCxhRcY7}

\bibitem{huang2017densely}
Huang, G., Liu, Z., Van Der~Maaten, L., Weinberger, K.Q.: Densely connected convolutional networks. In: Proceedings of the IEEE conference on computer vision and pattern recognition. pp. 4700--4708 (2017)

\bibitem{huang2021importance}
Huang, R., Geng, A., Li, Y.: On the importance of gradients for detecting distributional shifts in the wild. Advances in Neural Information Processing Systems  \textbf{34},  677--689 (2021)

\bibitem{huang2021training}
Huang, Y., Zhang, H., Shi, Y., Kolter, J.Z., Anandkumar, A.: Training certifiably robust neural networks with efficient local lipschitz bounds. Advances in Neural Information Processing Systems  \textbf{34},  22745--22757 (2021)

\bibitem{jiang2023diverse}
Jiang, W., Cheng, H., Chen, M., Wang, C., Wei, H.: Dos: Diverse outlier sampling for out-of-distribution detection. arXiv preprint arXiv:2306.02031  (2023)

\bibitem{jordan2020exactly}
Jordan, M., Dimakis, A.G.: Exactly computing the local lipschitz constant of relu networks. Advances in Neural Information Processing Systems  \textbf{33},  7344--7353 (2020)

\bibitem{kong2021opengan}
Kong, S., Ramanan, D.: Opengan: Open-set recognition via open data generation. In: Proceedings of the IEEE/CVF International Conference on Computer Vision. pp. 813--822 (2021)

\bibitem{krizhevsky2009learning}
Krizhevsky, A., Hinton, G., et~al.: Learning multiple layers of features from tiny images  (2009)

\bibitem{li2023rethinking}
Li, J., Chen, P., He, Z., Yu, S., Liu, S., Jia, J.: Rethinking out-of-distribution (ood) detection: Masked image modeling is all you need. In: Proceedings of the IEEE/CVF Conference on Computer Vision and Pattern Recognition. pp. 11578--11589 (2023)

\bibitem{li2023data}
Li, T., Qiao, F., Ma, M., Peng, X.: Are data-driven explanations robust against out-of-distribution data? In: Proceedings of the IEEE/CVF Conference on Computer Vision and Pattern Recognition. pp. 3821--3831 (2023)

\bibitem{li2020background}
Li, Y., Vasconcelos, N.: Background data resampling for outlier-aware classification. In: Proceedings of the IEEE/CVF Conference on Computer Vision and Pattern Recognition. pp. 13218--13227 (2020)

\bibitem{liang2017enhancing}
Liang, S., Li, Y., Srikant, R.: Enhancing the reliability of out-of-distribution image detection in neural networks. arXiv preprint arXiv:1706.02690  (2017)

\bibitem{lin2021mood}
Lin, Z., Roy, S.D., Li, Y.: Mood: Multi-level out-of-distribution detection. In: Proceedings of the IEEE/CVF conference on Computer Vision and Pattern Recognition. pp. 15313--15323 (2021)

\bibitem{liu2020energy}
Liu, W., Wang, X., Owens, J., Li, Y.: Energy-based out-of-distribution detection. Advances in neural information processing systems  \textbf{33},  21464--21475 (2020)

\bibitem{liu2023gen}
Liu, X., Lochman, Y., Zach, C.: Gen: Pushing the limits of softmax-based out-of-distribution detection. In: Proceedings of the IEEE/CVF Conference on Computer Vision and Pattern Recognition. pp. 23946--23955 (2023)

\bibitem{lloyd1982least}
Lloyd, S.: Least squares quantization in pcm. IEEE transactions on information theory  \textbf{28}(2),  129--137 (1982)

\bibitem{lo2022adversarially}
Lo, S.Y., Oza, P., Patel, V.M.: Adversarially robust one-class novelty detection. IEEE Transactions on Pattern Analysis and Machine Intelligence  \textbf{45}(4),  4167--4179 (2022)

\bibitem{losch2023certified}
Losch, M., Stutz, D., Schiele, B., Fritz, M.: Certified robust models with slack control and large lipschitz constants. arXiv preprint arXiv:2309.06166  (2023)

\bibitem{lu2024learning}
Lu, H., Gong, D., Wang, S., Xue, J., Yao, L., Moore, K.: Learning with mixture of prototypes for out-of-distribution detection. arXiv preprint arXiv:2402.02653  (2024)

\bibitem{ming2022delving}
Ming, Y., Cai, Z., Gu, J., Sun, Y., Li, W., Li, Y.: Delving into out-of-distribution detection with vision-language representations. Advances in Neural Information Processing Systems  \textbf{35},  35087--35102 (2022)

\bibitem{ming2022poem}
Ming, Y., Fan, Y., Li, Y.: Poem: Out-of-distribution detection with posterior sampling. In: International Conference on Machine Learning. pp. 15650--15665. PMLR (2022)

\bibitem{moon2022difficulty}
Moon, W., Park, J., Seong, H.S., Cho, C.H., Heo, J.P.: Difficulty-aware simulator for open set recognition. In: European Conference on Computer Vision. pp. 365--381. Springer (2022)

\bibitem{neal2018open}
Neal, L., Olson, M., Fern, X., Wong, W.K., Li, F.: Open set learning with counterfactual images. In: Proceedings of the European Conference on Computer Vision (ECCV). pp. 613--628 (2018)

\bibitem{netzer2011reading}
Netzer, Y., Wang, T., Coates, A., Bissacco, A., Wu, B., Ng, A.Y.: Reading digits in natural images with unsupervised feature learning  (2011)

\bibitem{nguyen2015deep}
Nguyen, A., Yosinski, J., Clune, J.: Deep neural networks are easily fooled: High confidence predictions for unrecognizable images. In: Proceedings of the IEEE conference on computer vision and pattern recognition. pp. 427--436 (2015)

\bibitem{noh2023simple}
Noh, S., Jeong, D., Lee, J.H.: Simple and effective out-of-distribution detection via cosine-based softmax loss. In: Proceedings of the IEEE/CVF International Conference on Computer Vision. pp. 16560--16569 (2023)

\bibitem{olber2023detection}
Olber, B., Radlak, K., Popowicz, A., Szczepankiewicz, M., Chachu{\l}a, K.: Detection of out-of-distribution samples using binary neuron activation patterns. In: Proceedings of the IEEE/CVF Conference on Computer Vision and Pattern Recognition. pp. 3378--3387 (2023)

\bibitem{oza2019c2ae}
Oza, P., Patel, V.M.: C2ae: Class conditioned auto-encoder for open-set recognition. In: Proceedings of the IEEE/CVF Conference on Computer Vision and Pattern Recognition. pp. 2307--2316 (2019)

\bibitem{park2023nearest}
Park, J., Jung, Y.G., Teoh, A.B.J.: Nearest neighbor guidance for out-of-distribution detection. In: Proceedings of the IEEE/CVF International Conference on Computer Vision. pp. 1686--1695 (2023)

\bibitem{park2022understanding}
Park, J., Park, H., Jeong, E., Teoh, A.B.J.: Understanding open-set recognition by jacobian norm of representation. arXiv preprint arXiv:2209.11436  (2022)

\bibitem{peng2024conjnorm}
Peng, B., Luo, Y., Zhang, Y., Li, Y., Fang, Z.: Conjnorm: Tractable density estimation for out-of-distribution detection. arXiv preprint arXiv:2402.17888  (2024)

\bibitem{perini2024unsupervised}
Perini, L., Davis, J.: Unsupervised anomaly detection with rejection. Advances in Neural Information Processing Systems  \textbf{36} (2024)

\bibitem{safaei2023open}
Safaei, B., Vibashan, V., de~Melo, C.M., Hu, S., Patel, V.M.: Open-set automatic target recognition. In: ICASSP 2023-2023 IEEE International Conference on Acoustics, Speech and Signal Processing (ICASSP). pp.~1--5. IEEE (2023)

\bibitem{shapley1953value}
Shapley, L.S., et~al.: A value for n-person games  (1953)

\bibitem{srinivas2022efficient}
Srinivas, S., Matoba, K., Lakkaraju, H., Fleuret, F.: Efficient training of low-curvature neural networks. Advances in Neural Information Processing Systems  \textbf{35},  25951--25964 (2022)

\bibitem{sun2021react}
Sun, Y., Guo, C., Li, Y.: React: Out-of-distribution detection with rectified activations. Advances in Neural Information Processing Systems  \textbf{34},  144--157 (2021)

\bibitem{sun2022dice}
Sun, Y., Li, Y.: Dice: Leveraging sparsification for out-of-distribution detection. In: European Conference on Computer Vision. pp. 691--708. Springer (2022)

\bibitem{tang2024cores}
Tang, K., Hou, C., Peng, W., Chen, R., Zhu, P., Wang, W., Tian, Z.: Cores: Convolutional response-based score for out-of-distribution detection. In: Proceedings of the IEEE/CVF Conference on Computer Vision and Pattern Recognition. pp. 10916--10925 (2024)

\bibitem{van2018inaturalist}
Van~Horn, G., Mac~Aodha, O., Song, Y., Cui, Y., Sun, C., Shepard, A., Adam, H., Perona, P., Belongie, S.: The inaturalist species classification and detection dataset. In: Proceedings of the IEEE conference on computer vision and pattern recognition. pp. 8769--8778 (2018)

\bibitem{wang2021review}
Wang, L., Han, M., Li, X., Zhang, N., Cheng, H.: Review of classification methods on unbalanced data sets. IEEE Access  \textbf{9},  64606--64628 (2021)

\bibitem{wang2024learning}
Wang, Q., Fang, Z., Zhang, Y., Liu, F., Li, Y., Han, B.: Learning to augment distributions for out-of-distribution detection. Advances in Neural Information Processing Systems  \textbf{36} (2024)

\bibitem{wang2023out}
Wang, Q., Ye, J., Liu, F., Dai, Q., Kalander, M., Liu, T., Hao, J., Han, B.: Out-of-distribution detection with implicit outlier transformation. arXiv preprint arXiv:2303.05033  (2023)

\bibitem{xiao2010sun}
Xiao, J., Hays, J., Ehinger, K.A., Oliva, A., Torralba, A.: Sun database: Large-scale scene recognition from abbey to zoo. In: 2010 IEEE computer society conference on computer vision and pattern recognition. pp. 3485--3492. IEEE (2010)

\bibitem{xu2023vra}
Xu, M., Lian, Z., Liu, B., Tao, J.: Vra: variational rectified activation for out-of-distribution detection. Advances in Neural Information Processing Systems  \textbf{36},  28941--28959 (2023)

\bibitem{xu2015turkergaze}
Xu, P., Ehinger, K.A., Zhang, Y., Finkelstein, A., Kulkarni, S.R., Xiao, J.: Turkergaze: Crowdsourcing saliency with webcam based eye tracking. arXiv preprint arXiv:1504.06755  (2015)

\bibitem{yang2021semantically}
Yang, J., Wang, H., Feng, L., Yan, X., Zheng, H., Zhang, W., Liu, Z.: Semantically coherent out-of-distribution detection. In: Proceedings of the IEEE/CVF International Conference on Computer Vision. pp. 8301--8309 (2021)

\bibitem{yang2024follow}
Yang, Y., Lee, K., Dariush, B., Cao, Y., Lo, S.Y.: Follow the rules: Reasoning for video anomaly detection with large language models. In: European Conference on Computer Vision (2024)

\bibitem{yu2015lsun}
Yu, F., Seff, A., Zhang, Y., Song, S., Funkhouser, T., Xiao, J.: Lsun: Construction of a large-scale image dataset using deep learning with humans in the loop. arXiv preprint arXiv:1506.03365  (2015)

\bibitem{yu2019unsupervised}
Yu, Q., Aizawa, K.: Unsupervised out-of-distribution detection by maximum classifier discrepancy. In: Proceedings of the IEEE/CVF international conference on computer vision. pp. 9518--9526 (2019)

\bibitem{yuan2024discriminability}
Yuan, Y., He, R., Dong, Y., Han, Z., Yin, Y.: Discriminability-driven channel selection for out-of-distribution detection. In: Proceedings of the IEEE/CVF Conference on Computer Vision and Pattern Recognition. pp. 26171--26180 (2024)

\bibitem{zagoruyko2016wide}
Zagoruyko, S., Komodakis, N.: Wide residual networks. arXiv preprint arXiv:1605.07146  (2016)

\bibitem{zhang2023mixture}
Zhang, J., Inkawhich, N., Linderman, R., Chen, Y., Li, H.: Mixture outlier exposure: Towards out-of-distribution detection in fine-grained environments. In: Proceedings of the IEEE/CVF Winter Conference on Applications of Computer Vision. pp. 5531--5540 (2023)

\bibitem{zhang2023openood}
Zhang, J., Yang, J., Wang, P., Wang, H., Lin, Y., Zhang, H., Sun, Y., Du, X., Zhou, K., Zhang, W., Li, Y., Liu, Z., Chen, Y., Li, H.: Openood v1.5: Enhanced benchmark for out-of-distribution detection. arXiv:2306.09301  (2023)

\bibitem{Zhang_2023_CVPR}
Zhang, Z., Xiang, X.: Decoupling maxlogit for out-of-distribution detection. In: Proceedings of the IEEE/CVF Conference on Computer Vision and Pattern Recognition (CVPR). pp. 3388--3397 (June 2023)

\bibitem{zheng2023out}
Zheng, H., Wang, Q., Fang, Z., Xia, X., Liu, F., Liu, T., Han, B.: Out-of-distribution detection learning with unreliable out-of-distribution sources. Advances in Neural Information Processing Systems  \textbf{36},  72110--72123 (2023)

\bibitem{zhou2017places}
Zhou, B., Lapedriza, A., Khosla, A., Oliva, A., Torralba, A.: Places: A 10 million image database for scene recognition. IEEE transactions on pattern analysis and machine intelligence  \textbf{40}(6),  1452--1464 (2017)

\bibitem{Zhu_2023_CVPR}
Zhu, F., Cheng, Z., Zhang, X.Y., Liu, C.L.: Openmix: Exploring outlier samples for misclassification detection. In: Proceedings of the IEEE/CVF Conference on Computer Vision and Pattern Recognition (CVPR). pp. 12074--12083 (June 2023)

\bibitem{zhu2024diversified}
Zhu, J., Geng, Y., Yao, J., Liu, T., Niu, G., Sugiyama, M., Han, B.: Diversified outlier exposure for out-of-distribution detection via informative extrapolation. Advances in Neural Information Processing Systems  \textbf{36} (2024)

\end{thebibliography}
\clearpage

\section{Supplementary Materials}
In the supplementary materials, we provide detailed implementation details and experimental results for both CIFAR and ImageNet benchmarks. 
Furthermore, we present an analysis to show the effectiveness of our OOD
sampling strategy. We then analyze the energy distribution of the ID and OOD samples. 

\subsection{Additional Implementation Details}
For the experiments on CIFAR we use SGD with a momentum of 0.9, and a weight decay of 0.0001. 
For \emph{GReg} on CIFAR, similar to \cite{liu2020energy}, we decrease the learning rate following a cosine annealing strategy, with a maximum learning rate of $1$ and a minimum of $0.001$.
We use a batch size of 64 and 32 for the CIFAR and ImageNet experiments, respectively.\\
For both \emph{GReg} and \emph{GReg+} on the ImageNet dataset we perform fine-tuning of a pre-trained DenseNet-121 model (in contrast to the CIFAR benchmarks where we train \emph{GReg+} from scratch) using the ADAM optimizer with an initial learning rate of $10^{-4}$ and decrease the learning rate to $10^{-5}$ at epoch 10. We then run \emph{GReg} and \emph{GReg+} to reach epochs 20 and 15, respectively. The other hyperparameters are the same as the CIFAR benchmarks.

To reproduce the results of MSP, ODIN, and Energy, we used the codebase of Energy\footnote{
\hyperref[https://github.com/wetliu/energy-ood]{https://github.com/wetliu/energy\_ood}}.
For the case of ReAct and DICE, we utilize the codebase of DICE\footnote{
\hyperref[https://github.com/deeplearning-wisc/dice]{https://github.com/deeplearning-wisc/dice}
}, and to reproduce the results of LINe\footnote{
\hyperref[https://github.com/YongHyun-Ahn/LINe-Out-of-Distribution-Detection-by-Leveraging-Important-Neurons]{https://github.com/YongHyun-Ahn/LINe-Out-of-Distribution-Detection-by-Leveraging-Important-Neurons}
}, and DOS\footnote{
\hyperref[https://github.com/lygjwy/DOS]{https://github.com/lygjwy/DOS}
}, we utilize their corresponding codebases. 
In the spirit of fairness, we run their codes with multiple specifications similar to the original manuscript and report the best results in our tables.
Furthermore, experiments requiring training are conducted with 3 different seeds and we report the average values in the tables.
All other methods are run with their corresponding default parameters outlined in their manuscript. 

\subsection{Detailed CIFAR-10/100 Benchmark Results}
    \Cref{tab:cifar10-full} and \Cref{tab:cifar100-full} show the complete and detailed performance of various OOD detection approaches on each of the six OOD test datasets. 
    \Cref{tab:cifar10-sample} compares the sampling methods on the CIFAR-10 dataset. As it can be seen, especially on the WRN and DenseNet architectures, the performance of different methods is saturated.

\subsection{Distribution Analysis}
\begin{figure*}[t]
        \centering
        \includegraphics[width=\columnwidth,trim={4cm 1cm 4cm 1cm}]{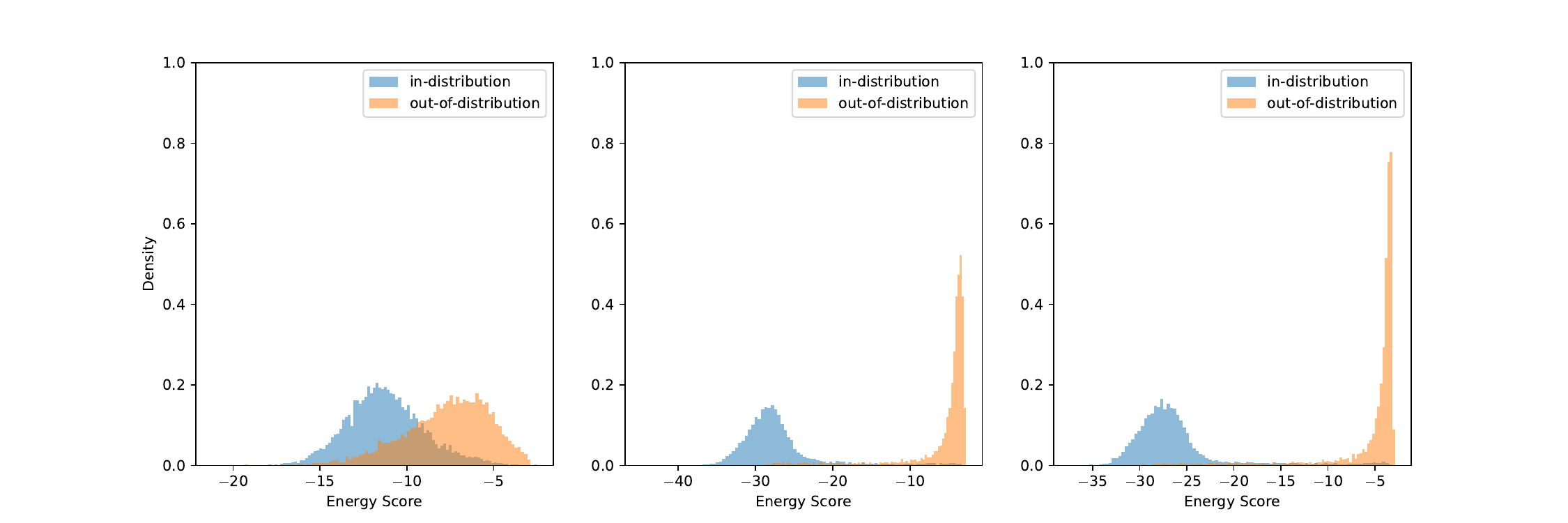} 
        \caption{Distributions of energy scores. The left figure shows the distribution of the pre-trained model, the middle shows the same for Energy loss and the right figure shows the distribution for \emph{GReg}.}
        \label{fig:energyDist}
    \end{figure*}

To further study the effectiveness of our method, we perform the following analysis.
    We choose a WRN model pre-trained on CIFAR-10 and fine-tune it with the energy loss and \emph{GReg}. 
    Next, we plot the distribution of energy scores for ID (CIFAR-10 test) and OOD (SVHN) datasets and compare the results.
    As it can be seen from \Cref{fig:energyDist},  adding gradient regularization to the energy loss enables the network to better distinguish ID samples from the OOD since the distance between the two distributions has increased, i.e., the OOD data are assigned higher energy scores, resulting in further distanced score distributions.

\begin{table}[t!]
	\caption{\textbf{Ablation study on the number of clusters.} FPR95 and AUROC percentages reported on CIFAR benchmarks on DenseNet.}
	\vskip -0.07in
	\label{tab:ablationCluster}
	\setlength\tabcolsep{6pt}
	\centering
	\renewcommand{\arraystretch}{1}
	\scalebox{0.67}{
		\begin{tabular}{llccccccccc}
			\toprule
			\multirow{2}{*}{\textbf{Num. of Clusters}} & \multicolumn{2}{c}{\textbf{CIFAR-10}} && \multicolumn{2}{c}{\textbf{CIFAR-100}}\\
			\cmidrule(lr){2-3}  \cmidrule(lr){5-6}
			 & \textbf{FPR95} $\downarrow$ & \textbf{AUROC} $\uparrow$  &&\textbf{FPR95} $\downarrow$ & \textbf{AUROC} $\uparrow$  \\
			\midrule

			 No Clustering & \textbf{7.93} & \textbf{98.12}  && 56.29 &	87.35\\
			 16 & 9.88 & 97.68  && 58.97 &	82.88  \\
			 32 &  12.1 & 97.15  && 46.41 &	88.433   \\
			 64 &  11.27 & 97.54  && \textbf{30.55} & \textbf{93.38} \\
			\bottomrule
		\end{tabular}
	} 
\end{table}
\subsection{Sampling}
In our setting, we are presented with three main possible clustering spaces: the image space, the feature space, or the logit space.
The image space contains the most information but there are two main drawbacks to using it. 
Firstly, the image space is far too large and high-dimensional, typically in the order of tens of thousands.
Secondly, although all the information is in the image, no processing has been done on the image and the features have not been extracted, meaning that in effect, there is also a lot of noise irrelevant to our task. 
The logit space is not suitable from the opposite perspective, i.e., most of the information has been stripped and only label-relevant information remains.
The feature space presents a sweet spot;  reasonable dimensionality, relevant features, and smaller levels of noise.
Consequently, we choose to cluster the samples in the \emph{feature space}. 

Another parameter of interest in the experiments is the number of clusters. 
In \Cref{tab:ablationCluster} we examine the effect of the number of clusters on the performance of our method on both CIFAR benchmarks using the DenseNet architecture. 
It can be observed that our sampling significantly improves the results on the CIFAR-100 benchmark, but does not have the same effect on CIFAR-10.
Our intuition suggests that one reason behind this observation might be that the CIFAR-10 benchmark is so simple that the DenseNet model can observe all the auxiliary data during the training phase and learn the information. 
However, in the case of CIFAR-100, as the benchmark is harder, sampling the data helps the model extract more informative samples. Note that the perplexity in comparing CIFAR-100 vs CIFAR-10 is that the datasets contain the same number of samples (60000 total images in both) but CIFAR-100 has 10 times the number of classes of CIFAR-10.
This can also be seen in our ImageNet experiments (see \Cref{tab:imagenet}). That is, as the task is harder, sampling using \emph{GReg+} is more effective and improves upon \emph{GReg} by large margins of $12\%$ and $2\%$ on  FPR and AUROC, respectively.

\subsection{Near-OOD vs Far-OOD Experiments}
Apart from the previous experiments, another way to categorize the OOD experiments is by grouping them into Near-OOD (hard-OOD) and Far-OOD (easy-OOD). 
The intuition behind this grouping is that if the ID data is close to the OOD data, it will be much harder to distinguish between them. Therefore, a powerful OOD method should perform well in both Near-OOD and Far-OOD benchmarks.

To evaluate our performance on the Near-OOD benchmarks, we followed the setup of \cite{zhang2023openood}. 
In Table \ref{tab:nearOOD}, we set CIFAR10 as ID and report the metrics on Near-OOD benchmarks (CIFAR100 and Tiny ImageNet) using a ResNet architecture.
On average, \emph{GReg+} improves the FPR of DOS by $4\%$ with comparable results in terms of AUROC.

To further compare our method with more auxiliary-based baselines, we use the setup of \cite{zhang2023openood} to perform additional experiments.
Table \ref{tab:Benchmarks}, compares our method versus three new benchmarks (MCD \cite{yu2019unsupervised}, UDG \cite{yang2021semantically}, MixOE\cite{zhang2023mixture}) that use auxiliary data.
% It can be seen that o
On average, \emph{GReg+} improves the FPR on CIFAR10 and CIFAR100 by $2.5 \%$ and $2.4 \%$, respectively, and achieves comparable AUROC. 
Note that, unlike DOS, \emph{GReg+} achieves these results with minimal decrease in ID accuracy. 
Furthermore, \emph{GReg+} outperforms the new benchmarks by large margins.

\begin{table}[t]
    \setlength{\tabcolsep}{2pt}
    \centering
    \resizebox{.7\columnwidth}{!}
    {%
        \begin{tabular}{c  c  c  c  c  c  c  c} 
        \hline
           \multirow{2}{*}{\textbf{ }} & \multirow{2}{*}{\textbf{Method}} & \multicolumn{2}{c}{\textbf{CIFAR-100}} & \multicolumn{2}{c}{\textbf{TIN}} & \multicolumn{2}{c}{\textbf{Average}}\\
           \cmidrule{3-8}
           & & FPR95 $\downarrow$  & AUROC $\uparrow$ & FPR95 $\downarrow$ & AUROC $\uparrow$ & FPR95 $\downarrow$ & AUROC $\uparrow$\\
           \cmidrule{2-8}
           \multirow{7}{*}{\rotatebox{90}{\textbf{CIFAR-10}}} 
           & ReAct \cite{sun2021react} & 75.5 & 85.2 & 67.61 & 87.70 & 71.56 & 86.47\\
           & Energy score \cite{liu2020energy} & 72.69 & 85.55 & 62.41 & 88.31 & 67.55 & 86.93\\
           & Dice \cite{sun2022dice} & 84.29 & 76.05 & 75.97 & 79.53 & 80.13 & 77.79\\
           \cmidrule{2-8}
           & Energy Loss \cite{liu2020energy} & 48.11 & 88.24 & 36.33 & 91.86 & 42.22 & 90.05 \\
           & DOS \cite{jiang2023diverse} & 29.24 & \textbf{93.38} & 16.08 & \textbf{96.16} & 22.66 & \textbf{94.77} \\
           % & \emph{GReg} & 44.97 & 89.73 & 30.07 & 93.75 & 37.52 & 91.74 \\
           & \emph{GReg+} & \textbf{23.87} & \textbf{93.23} & \textbf{13.12} & 95.32 & \textbf{18.49} & \textbf{94.28}\\
         \hline
        \end{tabular}%
    }
    \caption{Near-OOD comparison with CIFAR-10 as ID.}
    \label{tab:nearOOD}
\end{table}

\begin{table}[t]
    \setlength{\tabcolsep}{2pt}
    \centering
    \resizebox{.8\columnwidth}{!}
    {%
        \begin{tabular}{c  c  c  c  c  c c c c} 
        \hline
           \multirow{2}{*}{\textbf{}} & \multirow{2}{*}{\textbf{Method}} & \multicolumn{2}{c}{\textbf{Near-OOD}} & \multicolumn{2}{c}{\textbf{Far-OOD}} & \multicolumn{2}{c}{\textbf{Average}} & \multirow{2}{*}{\textbf{ID Acc}}\\
           \cmidrule{3-8}
           & & FPR95 $\downarrow$  & AUROC $\uparrow$ & FPR95 $\downarrow$ & AUROC $\uparrow$ & FPR95 $\downarrow$ & AUROC $\uparrow$ & \\
           \cmidrule{2-9}
           \multirow{5}{*}{\rotatebox{90}{\textbf{CIFAR-10}}}
           & MCD \cite{yu2019unsupervised} & 30.17 & 91.03  & 32.03 & 91.00 & 31.1 & 91.0 & 94.9  \\
           & UDG \cite{yang2021semantically} & 35.34  & 89.91 & 20.35 & 94.06 & 27.8 & 91.9 & 92.3  \\
           & MixOE \cite{zhang2023mixture} & 51.45 & 88.73 & 33.84 & 91.93 &  42.6 & 90.3 & 94.5  \\
           & DOS \cite{jiang2023diverse} & 22.66 & 94.77 & 8.01 & 97.75 & 15.3 & \textbf{96.2} & 78.5 \\
           & \emph{GReg+} & 18.49 & 94.28 & 7.47 & 96.72 & \textbf{12.9} & 95.5 & 91.1 \\ 
           \cmidrule{2-9}
           \multirow{5}{*}{\rotatebox{90}{\textbf{CIFAR-100}}} 
           & MCD \cite{yu2019unsupervised} & 55.88 & 77.07 & 54.39 & 74.72 & 55.1 & 75.8 & 75.8\\
           & UDG \cite{yang2021semantically} & 61.42 & 78.02  & 59.00 & 79.59 & 60.2 & 78.8 & 71.5\\
           & MixOE \cite{zhang2023mixture} & 55.22 & 80.95 & 63.88 & 76.40 & 59.5 & 78.6 & 75.3 \\
           & DOS \cite{jiang2023diverse}& 56.33 & 79.63 & 35.52 & 87.73 & 45.9 & 83.6 & 47.7\\
           & \emph{GReg+} & 48.26 & 82.50 & 38.79 & 86.4 & \textbf{43.5} & \textbf{84.5} & 72.4\\
         \hline
        \end{tabular}%
    }
    \caption{Additional Experiments on Near-OOD and Far-OOD.}
    \label{tab:Benchmarks}
\end{table}

\begin{table*}[t!]
\centering
\caption{\textbf{Comparison on the CIFAR-10 benchmark.} The experimental results are reported over three trials. AUROC and FPR95 are percentages.} 
\label{tab:cifar10-full}
\scalebox{0.5}{
\begin{tabular}{ccccccccccccccccccc}
    \toprule
  \multirow{3}{*}{\textbf{Network}} & \multirow{3}{*}{\textbf{Method}}  & \multicolumn{12}{c}{\textbf{OOD Datasets}} & 
  \multicolumn{2}{c}{\multirow{3}{*}{\textbf{Average}}}\\ 
  \cline{3-14}\\
 & \multicolumn{1}{c}{} & \multicolumn{2}{c}{\textbf{Textures}} & \multicolumn{2}{c}{\textbf{SVHN}} & \multicolumn{2}{c}{\textbf{Places}} & \multicolumn{2}{c}{\textbf{LSUN-c}} & \multicolumn{2}{c}{\textbf{LSUN-r}} &  \multicolumn{2}{c}{\textbf{iSUN}} &  &&\\
 & \multicolumn{1}{c}{} & FPR95 $\downarrow$  & AUROC $\uparrow$ & FPR95 $\downarrow$ & AUROC $\uparrow$ & FPR95 $\downarrow$ & AUROC $\uparrow$ & FPR95 $\downarrow$ & AUROC $\uparrow$ & FPR95 $\downarrow$ & AUROC $\uparrow$
 & FPR95 $\downarrow$ & AUROC $\uparrow$ & FPR95 $\downarrow$ & AUROC $\uparrow$\\
 \hline
\multirow{10}{*}{ResNet} 
& MSP \cite{hendrycks2017a} & 58.78	& 90.03 & 73.02 & 89.42 & 61.56 & 87.98 & 43.88 &	94.21 &	53.31 &	91.38 &	57.73 &	89.95 &	58.04 &	90.49 \\
& ODIN \cite{liang2017enhancing} &50.42 & 86.89 &	40.27 &	91.84 &	48.65 &	86.56 &	7.81 &	98.21 &	26.48 &	94 & 33.14 & 92.06 & 34.46 & 91.59 \\
& Energy score\cite{liu2020energy} & 53.26 & 89.12 & 58.36 & 91.01 & 44.51 &	89.45 &	12.57 &	97.64 &	30.48 &	94.11 &	36.77 &	92.16 &	39.32 &	92.24\\
& ReAct \cite{sun2021react} &52.04 & 89.68 & 60.2 &	90.46 &	44.68 &	89.26 &	13.36 &	97.51 &	29.95 &	94.29 &	36.43 &	92.63 &	39.44 &	92.30 \\
& DICE\cite{sun2022dice} & 54.73 &	88.34 &	59.4 &	89.81 &	44.09 &	89.39 &	22.38 &	95.72 &	33.32 &	93.17 &	40.52 &	91.08 &	42.40 &	91.25\\
& LINe\cite{ahn2023line} & 56.97 &	88.07 &	66.57 &	87.4 &	46.05 &	88.62 &	24.9 & 95.13 & 34.66 & 93.12 & 42.38 & 91.21 & 45.25 & 90.59\\
\cline{2-17}
&OE\cite{hendrycks2018deep} & 19.46 & 95.98 & 13.94 & 97.22 & 38.38 & 90.76 &	8.15 &	98.20 &	21.03 &	95.87 &	18.58 &	96.26 &	19.92 &	95.71\\
&Energy Loss \cite{liu2020energy} & 12.95 &	97.17 &	4.22 &	98.89 &	27.31 &	94.05 &	5.47 &	98.69 &	7.87 &	98.24 &	9.04 &	98.16 &	11.14 &  97.53\\
& OpenMix \cite{Zhu_2023_CVPR} & 17.76 & 96.77 &	41.12 &	94.44 &	27.68 &	94.4 &	8.33 &	98.32 & 19.06 &	96.87 &	19.54 &	96.79 & 22.24 & 96.26\\
\cline{2-17}
& GReg & 5.96 &	98.36 &	3 &	98.83 &	22.76 &	94.8 &	3.03 &	99 & 6.5 &	98.4 &	6.2 &	98.33 & 7.91 & 97.95\\
\midrule

\multirow{10}{*}{WRN} 
& MSP \cite{hendrycks2017a} & 59.53 & 88.45 & 48.53 & 91.74 & 59.86 & 88.29 &	31.15 &	95.6 &	53.22 &	91.14 &	56.87 &	89.58 &	51.52 &	90.8 \\
& ODIN \cite{liang2017enhancing} &54.52 & 80.45 & 46.6 & 85.68 & 48.57 & 86.04 & 10.19 &	97.84 &	22.86 &	95.05 &	28.64 &	93.87 &	35.23 &	89.82 \\
& Energy score\cite{liu2020energy} &52.52 &	85.38 &	36.58 &	90.73 &	39.88 &	89.87 &	8.2 & 98.34 & 28.8 & 93.87 & 34.47 & 92.38 & 33.40 & 91.76\\
& ReAct \cite{sun2021react} &53.47 & 86.58 &	41.46 &	89.34 &	41 &	90.51 &	13.21 &	97.4 &	34.95 &	93.30 &	41.15 &	91.82 &	37.54 &	91.49 \\
& DICE\cite{sun2022dice} & 58.97 &	84.24 &	46.09 &	88.22 &	42.55 &	89.69 &	2.87 & 99.27 & 23.6 & 95.29 & 30.66 & 93.78 & 34.12 & 91.74\\
& LINe\cite{ahn2023line} & 57.13 &	83.81 &	50.2 &	83.11 &	47.35 &	87.67 &	2 &	99.47 &	27.32 &	94.32 &	33.65 &	92.82 &	36.27 & 90.2\\
\cline{2-17}
&OE\cite{hendrycks2018deep} & 22.56 & 95.24 & 27.41 & 95.13 & 33.65 & 92.24 &	8.88 &	98.19 &	17.58 &	96.22 &	16.68 &	96.27 &	21.12 &	95.55\\
&Energy Loss \cite{liu2020energy} &11.47 &	97.3 &	23.93 &	94.93 &	22.6 &	95.37 &	5.07 &	98.83 &	8.07 &	98.17 &	7.53 &	98.27 &	13.11 &	97.14\\
& OpenMix \cite{Zhu_2023_CVPR} &21.17 &	95.85 &	26.63 &	95.65 &	27.17 &	94.13 &	14.32 &	97.35 &	21.8 &	96.42 &	20.45 &	96.61 &	21.92 &	96\\
\cline{2-17}
& GReg &7.83 &	98.16 &	6.23 &	98.4 & 19.9 & 95.66 &	3.8 &	98.93 &	5.26 &	98.7 &	4.66 &	98.76 &	7.95 &	98.10\\
\midrule

\multirow{10}{*}{DenseNet} 
& MSP \cite{hendrycks2017a} &66.3 &	87.09 &	44.64 &	93.86 &	63.16 &	88.35 &	43.34 &	94.17 &	48.9 & 93.4 & 49.05 & 93.35 & 52.56 & 91.70 \\
& ODIN \cite{liang2017enhancing} & 55.62 &	85.03 &	29.03 &	94.86 &	42.44 &	91.11 &	11.94 &	97.67 &	4.86 &	98.96 &	5.39 &	98.9 &	24.88 &	94.42 \\
& Energy score\cite{liu2020energy} & 60.03 & 85.17 & 35.2 &	94.76 &	40.9 &	91.58 &	9.5 & 98.17 & 13.78 & 97.51 & 14.57 & 97.4 & 28.99 & 94.09\\
& ReAct \cite{sun2021react} &50.47 & 90.53 & 29.81 & 95.56 & 40.35 & 92.05 &	10.04 &	98.11 &	11.44 &	97.79 &	12.92 &	97.62 &	25.83 &	95.27 \\
& DICE\cite{sun2022dice} &56.26 & 86.42 & 40.47 & 93.99 & 39.6 & 91.82 &	3.79 & 99.15 & 8.71 & 98.19 & 9.42 &	98.11 &	26.37 &	94.61\\
& LINe\cite{ahn2023line} & 23.35 &	95.11 &	12.22 &	97.56 &	43.72 &	91.13 &	0.61 &	99.83 &	4.09 &	99.1 &	5.06 &	99.02 &	14.84 &	96.95\\
\cline{2-17}
&OE\cite{hendrycks2018deep} & 20.93 & 96.04 & 12.19 & 97.68 & 39.18 & 91.69 &	7.15 &	98.57 &	25.45 &	95.42 &	25.72 &	95.41 &	21.77 &	95.8\\
&Energy Loss \cite{liu2020energy} &13.43 &	97.04 &	20.08 &	95.52 &	20.09 &	95.41 &	3.53 & 99.19 &	4.55 &	98.84 &	5.87 & 	98.63 &	11.26 &	97.44\\
& OpenMix \cite{Zhu_2023_CVPR} & 24.8 &	95.3 &	40.59 &	92.58 &	29.13 &	93.88 &	12.39 &	97.64 &	15.4 &	97.23 &	14.9 &	97.31 &	22.87 &	95.66\\
\cline{2-17}
& GReg & 8.3 &	97.9 &	3.7 & 98.83 & 20.13 & 95.66 & 3.1 &	99.23 &	6.36 &	98.53 & 6.03 &	98.56 &	7.93 &	98.12\\
\bottomrule
\end{tabular}
}
\end{table*}

\begin{table*}[t!]
\centering
\caption{\textbf{Comparison on the CIFAR-100 benchmark.} The experimental results are reported over three trials. AUROC and FPR95 are percentages.} 
\label{tab:cifar100-full}
\scalebox{0.50}{
\begin{tabular}{lcccccccccccccccccc}
    \toprule
  \multirow{3}{*}{\textbf{Network}} & \multirow{3}{*}{\textbf{Method}}  & \multicolumn{12}{c}{\textbf{OOD Datasets}} & 
  \multicolumn{2}{c}{\multirow{3}{*}{\textbf{Average}}}\\ 
  \cline{3-14}\\
 & \multicolumn{1}{c}{} & \multicolumn{2}{c}{\textbf{Textures}} & \multicolumn{2}{c}{\textbf{SVHN}} & \multicolumn{2}{c}{\textbf{Places}} & \multicolumn{2}{c}{\textbf{LSUN-c}} & \multicolumn{2}{c}{\textbf{LSUN-r}} &  \multicolumn{2}{c}{\textbf{iSUN}} &  &&\\
 & \multicolumn{1}{c}{} & FPR95 $\downarrow$  & AUROC $\uparrow$ & FPR95 $\downarrow$ & AUROC $\uparrow$ & FPR95 $\downarrow$ & AUROC $\uparrow$ & FPR95 $\downarrow$ & AUROC $\uparrow$ & FPR95 $\downarrow$ & AUROC $\uparrow$
 & FPR95 $\downarrow$ & AUROC $\uparrow$ & FPR95 $\downarrow$ & AUROC $\uparrow$\\
 \hline
\multirow{13}{*}{ResNet} 
& MSP \cite{hendrycks2017a} &82.18 & 75.44 & 70.19 & 82.53 & 82.09 & 75.06 &	64.37 &	85.43 &	74.74 &	81.24 &	74.39 &	80.94 &	74.66 &	80.1 \\
& ODIN \cite{liang2017enhancing} & 76.44 &	77.59 &	65.24 &	84.61 &	83.37 &	74.62 &	41.63 &	93.07 &	59.85 &	88 & 57.48 & 88.09 & 64 & 84.33 \\
& Energy score\cite{liu2020energy} & 81.82 & 77.71 & 57.08 & 89.09 & 84.12 &	75.4 &	50.24 &	91.71 &	69.73 &	85.49 &	67.77 &	85.5 &	68.46 &	84.15\\
& ReAct \cite{sun2021react} & 67.69 &	85.5 &	48.94 &	91.8 &	82.29 &	77.2 &	31.54 &	94.61 &	63.9 &	88.53 &	61.96 &	87.95 &	59.38 &	87.59 \\
& DICE\cite{sun2022dice} & 88.92 &	71.5 &	69.78 &	85.12 &	87.81 &	72.13 &	59.78 &	89.08 &	77.69 &	82.87 &	79.81 &	81.42 &	77.29 &	80.35\\
& LINe\cite{ahn2023line} & 78.71 &	79.96 &	66.2 &	86.98 &	87.71 &	72.88 &	43.98 &	91.09 &	78.27 &	83.48 &	77.69 &	83.19 &	72.09 &	82.93\\
\cline{2-17}
&OE\cite{hendrycks2018deep} &73.91 & 73.14 & 86.45 & 70.18 & 69.28 & 77.75 &	44.44 &	86.51 &	94.55 &	38.80 &	95.30 &	36.98 &	77.32 &	63.89\\
&Energy Loss \cite{liu2020energy} & 57.3 &	83.9 &	75.06 &	84.33 &	82.2 &	75.8 & 35.03 & 93.93 & 62.06 & 83.4 & 65 & 82.93 & 62.77 & 84.05\\
&POEM \cite{ming2022poem} & 67.92 &	77.69 &	78.72 &	78.6 &	89.07 &	68.76 &	41.97 &	92.24 &	46.27 &	88.31 &	42.72 &	89.01 &	61.11 &	82.43\\
& OpenMix \cite{Zhu_2023_CVPR} & 59.87 & 87.54 & 72.91 & 86.76 & 75.73 &	80 & 48.75 &	91.31 &	48.02 &	92.25 &	52.56 &	90.71 &	59.64 &	88.09\\
& DOS \cite{jiang2023diverse} & 52.29 &	88.1 &	47.57 &	91.44 &	77.99 &	81.42 &	54.75 &	89 & 45.5 &	90.76 &	49.62 &	89.08 &	54.62 &	88.30\\
\cline{2-17}
& GReg & 58.73 &	83.73 &	42 & 91.7 &	78.4 &	76.7 &	28.56 &	94.6 &	74.56 &	75.2 &	75.33 &	75.63 &	59.6 &	82.92\\
& GReg+ & 48.74 & 88.2 & 42.2 &	92.45 &	69.61 &	83.35 &	43.86 &	90.75 &	46.04 &	89.94 &	54.22 &	87.8 & 50.78 &	88.75\\
\midrule

\multirow{13}{*}{WRN} 
& MSP \cite{hendrycks2017a} & 83.05 & 73.75 & 83.88 & 71.97 & 82.03 & 73.91 &	66.59 &	83.71 &	83.11 &	74.12 &	83.66 &	74.77 &	80.39 &	75.37\\
& ODIN \cite{liang2017enhancing} & 76.57 &	75.13 &	90.37 &	67.08 &	81.12 &	74.21 &	36.44 &	93.32 &	62 & 84.55 & 61.23 & 84.89 & 67.95 & 79.86\\
& Energy score\cite{liu2020energy} & 79.78 & 76.47 & 85.57 & 74.43 & 80.07 &	75.69 &	36.09 &	93.4 &	80.6 &	77.72 &	82.82 &	77.86 &	74.15 &	79.26\\
& ReAct \cite{sun2021react} & 67.12 & 82.97 &	74.45 &	88.28 &	81.21 &	74.24 &	36.63 &	91.67 &	81.23 &	72 & 83.31 & 72.07 & 70.65 & 80.20\\
& DICE\cite{sun2022dice} &85.32 &	72.84 &	87.21 &	71.39 &	81.02 &	74.95 &	16.73 &	96.83 &	79.86 &	81.64 &	83.82 &	80.08 &	72.32 &	79.62 \\
& LINe\cite{ahn2023line} & 67.34 & 81.98 &	81.73 &	83.6 &	84.03 &	71.03 &	16.93 &	96.61 &	77.98 &	76.26 &	77	& 76.37 & 67.50 &80.97\\
\cline{2-17}
&OE\cite{hendrycks2018deep} & 73.24 & 72.47 & 71.6 & 81.3 &	69.2 & 77.97 &	40.46 &	87.26 &	95.04 &	38.36 &	95.77 &	36.84 &	74.22 &	65.7\\
&Energy Loss \cite{liu2020energy} & 68.96 &	80.66 &	87.7 &	73.6 &	83.26 &	74.36 &	55.53 &	88.53 &	47.7 &	81.96 &	50.6 &	82 &65.62 &	80.18\\
&POEM \cite{ming2022poem} & 79.23 &	70.17 &	90.37 &	67.89 &	85.89 &	70.51 &	26.09 &	94.35 &	45.3 &	88.59 &	48.59 &	86.92 &	62.58 &	79.74\\
& OpenMix \cite{Zhu_2023_CVPR} & 64.91 & 83.61 & 87.25 & 70.34 & 73.17 &	78.14 &	59.18 &	87.77 &	75.19 &	78.46 &	79.02 &	77.05 &	73.12 &	79.23\\
& DOS \cite{jiang2023diverse} & 41.89 &	91.58 &	15.07 &	97.33 &	59.18 &	88.24 &	32.12 &	94.38 &	61.31 &	86.78 &	62 & 86.22 & 45.26 & 90.76\\
\cline{2-17}
& GReg & 54.5 &	85.53 &	57.33 &	89.76 &	74.36 &	79.53 &	43.86 &	91.73 &	58.33 &	86.8 &	61.16 &	86.03 &	58.26 &	86.56\\
& GReg+ & 51.6 & 90 & 25.73 & 95.96 & 65.16 & 87.56 & 41.68 & 93.42 &	51.61 &	89.96 &	52.93 &	89.24 &	48.12 &	91.02\\
\midrule

\multirow{13}{*}{DenseNet} 
& MSP \cite{hendrycks2017a} & 86.12 & 70.66 & 85.55 & 72.33 & 83.25 & 74.01 &	77.45 &	78 & 92.18 & 57.09 & 90.71 & 60.59 & 85.87 & 68.78\\
& ODIN \cite{liang2017enhancing} & 80.14 & 73.74 & 80.63 &	83.75 &	76.78 &	78.75 &	43.87 &	91.25 &	69.6 &	80.44 &	68.8 &	81.4 &	69.97 &	81.55\\
& Energy score\cite{liu2020energy} &84.81 &	70.29 &	88.6 &	80.99 &	78.12 &	77.48 &	51.08 &	88.73 &	84.56 &	69.31 &	86.3 &	69.96 &	78.91 &	76.12\\
& ReAct \cite{sun2021react} & 78.22 & 77.9 & 88.02 & 78.76 & 81.98 & 73.53 &	53.51 &	88.13 &	76.23 &	81.32 &	79.7 &	80.97 &	76.27 &	80.10\\
& DICE\cite{sun2022dice} & 84.34 & 71.02 & 87.51 & 81.86 &	78.19 &	78.15 &	14.75 &	97.44 &	75.36 &	77.77 &	78.7 &	76.8 &	69.80 &	80.50\\
& LINe\cite{ahn2023line} & 39.24 & 87.91 & 31.6 & 91.7 & 88.48 & 63.82 & 5.75 & 98.85 & 23.33 & 94.95 & 22.6 & 95.13 & 35.16 & 88.72\\
\cline{2-17}
&OE\cite{hendrycks2018deep} & 74.55 & 77.07 & 61.17 & 87.34 & 66.75 & 81.56 &	24.98 &	95.1 & 65.02 & 83.12 & 70.65 & 81.11 & 60.52 & 84.21\\
&Energy Loss \cite{liu2020energy} & 70.03 &	81.3 & 70.78 &	88.42 &	62.45 &	85.15 &	26.09 &	95.38 &	77.59 &	75.23 &	79.32 &	77.69 &	64.37 &	83.86\\
&POEM \cite{ming2022poem} & 75.48 &	73.04 &	83.57 &	71.49 &	83.6 &	73.99 &	33.96 &	93.43 &	39.74 &	83.58 &	39.61 &	83.33 &	59.33 &	79.81 \\
& OpenMix \cite{Zhu_2023_CVPR} & 63.66 & 84.05 & 72.27 & 85.77 & 73.17 &	80.19 &	48.79 &	90.8 &	71.76 &	83.77 &	74.1 &	82.23 &	67.29 &	84.47 \\
& DOS \cite{jiang2023diverse} & 38.3 & 91.72 &	11.57 &	97.88 &	57.06 &	88.91 &	25.32 &	95.58 &	38.43 &	92.75 &	38.85 &	92.49 &	34.92 &	93.22\\
\cline{2-17}
& GReg & 64.63 & 81.74 & 45.79 & 92.53 & 78.1 &	78.1 & 31.27 & 94.53 &	55.98 &	89.43 &	61.96 &	87.8 &	56.29 &	87.35\\
& GReg+ & 41.83 & 90.19 & 14.44 &	97.06 &	53.56 &	89.11 &	19.49 &	95.8 &	26.01 &	94.36 &	27.95 &	93.77 &	30.55 &	93.38\\
\bottomrule
\end{tabular}
}
\end{table*}

\begin{table*}[t!]
\centering
\caption{\textbf{Comparison of methods with sampling on the CIFAR-10 benchmark.} The experimental results are reported over three trials. AUROC and FPR95 are percentages.} 
\label{tab:cifar10-sample}
\scalebox{0.50}{
\begin{tabular}{lcccccccccccccccccc}
    \toprule
  \multirow{3}{*}{\textbf{Network}} & \multirow{3}{*}{\textbf{Method}}  & \multicolumn{12}{c}{\textbf{OOD Datasets}} & 
  \multicolumn{2}{c}{\multirow{3}{*}{\textbf{Average}}}\\ 
  \cline{3-14}\\
 & \multicolumn{1}{c}{} & \multicolumn{2}{c}{\textbf{Textures}} & \multicolumn{2}{c}{\textbf{SVHN}} & \multicolumn{2}{c}{\textbf{Places}} & \multicolumn{2}{c}{\textbf{LSUN-c}} & \multicolumn{2}{c}{\textbf{LSUN-r}} &  \multicolumn{2}{c}{\textbf{iSUN}} &  &&\\
 & \multicolumn{1}{c}{} & FPR95 $\downarrow$  & AUROC $\uparrow$ & FPR95 $\downarrow$ & AUROC $\uparrow$ & FPR95 $\downarrow$ & AUROC $\uparrow$ & FPR95 $\downarrow$ & AUROC $\uparrow$ & FPR95 $\downarrow$ & AUROC $\uparrow$
 & FPR95 $\downarrow$ & AUROC $\uparrow$ & FPR95 $\downarrow$ & AUROC $\uparrow$\\
 \hline
\multirow{3}{*}{ResNet} 
&POEM\cite{ming2022poem} & 33.24 &	93.51 &	58.95 &	89.58 &	61.82 &	86.26 &	62.78 &	90.13 &	11.22 &	97.96 &	10.53 &	98.03 &	39.76 &	92.58\\
&DOS\cite{jiang2023diverse} & 7.76 & 98.74 & 4.82 &	99.19 &	21.3 &	95.50 &	9.68 &	98.39 &	10.86 &	97.93 &	12.31 &	97.68 &	11.12 &	97.91\\
\cline{2-17}
& GReg & 5.96 &	98.36 &	3 &	98.83 &	22.76 &	94.8 &	3.03 &	99 & 6.5 &	98.4 &	6.2 &	98.33 & \textbf{7.91} & \textbf{97.95}\\
& GReg+ & 7.70 & 98.47 & 5.65 &	98.79 &	24.25 &	94.80 &	9.54 &	98.18 &	14.56 &	97.30 &	15.04 &	97.11 &	12.79 &	97.44\\
\midrule
\midrule

\multirow{3}{*}{WRN} 
&POEM\cite{ming2022poem} &  31.08 &	93.84 &	64.09 &	87.23 &	57.86 &	87.62 &	29.51 &	95.01 &	25.97 &	95.08 &	23.17 &	95.48 &	38.61 &	92.38\\
&DOS\cite{jiang2023diverse} & 4.33 &	99 &	1.78 & 99.32 & 10.89 & 97.16 &	3.73 &	98.98 &	9.18 &	98.2 &	9.86 &	98.17 &	6.63 &	\textbf{98.47}\\
\cline{2-17}
& GReg &7.83 &	98.16 &	6.23 &	98.4 & 19.9 & 95.66 &	3.8 &	98.93 &	5.26 &	98.7 &	4.66 &	98.76 &	7.95 &	98.10\\
& GReg+ & 4.19 & 98.52 & 2.92 &	98.59 &	11.38 &	96.71 &	3.88 &	98.63 &	5.63 &	98.47 &	6.57 &	98.27 &	\textbf{5.76} &	98.2\\
\midrule
\midrule\textbf{}

\multirow{3}{*}{DenseNet} 
&POEM\cite{ming2022poem} & 36.79 &	91.31 &	31.69 &	94.34 &	53.69 &	88.34 &	43.57 &	92.84 &	5.06 &	99.03 &	4.11 &	99.23 &	29.15 &	94.18\\
&DOS\cite{jiang2023diverse} & 2.61 & 99.39 & 0.67 &	99.65 &	7.16 &	97.90 &	0.88 & 99.56 & 1.61 & 99.21 & 1.53 &	99.27 &	\textbf{2.41} &	\textbf{99.16}\\
\cline{2-17}
& GReg & 8.3 &	97.9 &	3.7 & 98.83 & 20.13 & 95.66 & 3.1 &	99.23 &	6.36 &	98.53 & 6.03 &	98.56 &	7.93 &	98.12\\
& GReg+ & 6.21 & 98.29 & 2.19 &	98.73 &	21.5 &	95.1 &	6.87 &	98.27 &	15.53 &	97.45 &	15.36 &	97.43 &	11.27 &	97.54\\
\bottomrule
\end{tabular}
}
\end{table*}

\end{document}